\documentclass[]{article}

\usepackage{cite}
\usepackage{graphicx}
\usepackage{hyperref}
\usepackage{url}
\usepackage[table]{xcolor}  
\usepackage{multirow}       
\usepackage{adjustbox}      

\usepackage{amsmath}		
\usepackage{MnSymbol}		%
\usepackage{wasysym}		%

\usepackage[table]{xcolor}  
\usepackage{multirow}       
\usepackage{adjustbox}      
\usepackage{booktabs}       

\usepackage{tikz}           
\usepackage{pgfplots}
\usetikzlibrary{patterns}
\usepackage{pgfplots,lipsum}
\usepackage{enumerate}

\usepackage[affil-it]{authblk}%

\usepackage{subcaption}   
\usepackage[normalem]{ulem}	

\usepackage{algorithm}
\usepackage{algpseudocode}
\usepackage{algcompatible}
\algrenewcommand\algorithmicindent{2.0em}%

\algdef{SE}{MyFor}{EndMyFor}[1]{%
	\textbf{foreach} #1}{%
	\textbf{end foreach}}

\title{Knowledge Graph-Enabled Text-Based Automatic Personality Prediction}

\author[1]{Majid Ramezani\thanks{Corresponding author: m\_ramezani@tabrizu.ac.ir}}
\author[1]{Mohammad-Reza Feizi-Derakhshi\thanks{Corresponding author: mfeizi@tabrizu.ac.ir}}
\author[2]{Mohammad-Ali Balafar\thanks{balafarila@tabrizu.ac.ir}}

\affil[1]{Computerized Intelligence Systems Laboratory, Department of Computer Engineering, Faculty of Electrical and Computer Engineering, University of Tabriz, Tabriz, Iran}

\affil[2]{Department of Computer Engineering, Faculty of Electrical and Computer Engineering, University of Tabriz, Tabriz, Iran}

\date{}

\begin{document}

\maketitle

\begin{abstract}
How people think, feel, and behave primarily is a representation of their personality characteristics. By being conscious of the personality characteristics of individuals whom we are dealing with or deciding to deal with, one can competently ameliorate the relationship, regardless of its type. With the rise of Internet-based communication infrastructures (social networks, forums, etc.), a considerable amount of human communications take place there. The most prominent tool in such communications is the language in written and spoken form that adroitly encodes all those essential personality characteristics of individuals. Text-based Automatic Personality Prediction (APP) is the automated forecasting of the personality of individuals based on the generated/exchanged text contents. This paper presents a novel knowledge graph-enabled approach to text-based APP that relies on the Big Five personality traits. To this end, given a text, a knowledge graph which is a set of interlinked descriptions of concepts was built by matching the input text's concepts with DBpedia knowledge base entries. Then, due to achieving a more powerful representation, the graph was enriched with the DBpedia ontology, NRC Emotion Intensity Lexicon, and MRC psycholinguistic database information. Afterwards, the knowledge graph which is now a knowledgeable alternative for the input text was embedded to yield an embedding matrix. Finally, to perform personality predictions the resulting embedding matrix was fed to four suggested deep learning models independently, which are based on convolutional neural network (CNN), simple recurrent neural network (RNN), long short term memory (LSTM), and bidirectional long short term memory (BiLSTM). The results indicated considerable improvements in prediction accuracies in all of the suggested classifiers.

\end{abstract}


\section{Introduction}
\label{section-1}
\textit{Personality} is the enduring set of traits and styles that an individual exhibits; those characteristics that represent his/her dispositions, namely natural tendencies or personal inclinations \cite{BERGNER2020100759}. Being aware of the personality characteristics of people, will help them to improve their relationship management skills and also ameliorate their interpersonal communications; regardless of the type of relationship, as it happens between two friends, the boss and employee, investor and investee, seller and buyer, between members of a family and so on.

With the advent of social networks and their remarkable fortune among people, nowadays a great deal of communication happens through social networks. \textit{Language}, as the main communication tool among humans that competently represents their thoughts, emotions, opinions, and totally \textit{personality}, is also used in written and spoken form among social networks' users to communicate with each other. Admittedly, having some information about the personality with whom you are communicating, would be so advantageous. It can be carried out by analyzing the exchanged \textit{texts} (which is also known as \textit{written language}), among the users of such information infrastructures. Accordingly, the automatic prediction of human personality through computational approaches is called \textit{Automatic Personality Prediction} (\textit{APP}). 

What we know about \textit{text-based APP} is largely based upon empirical studies that have investigated how to exploit different methodologies for the purpose of personality prediction of individuals in Internet-based infrastructures (like social networks). Actually, various hypotheses regarding this issue can be found that they are commonly concerned about achieving a more \textit{knowledgeable} substitutions for text elements to deal with, rather than pure strings of characters. 

In the history of text-based APP, initial investigations have mostly focused on linguistic features of text elements to achieve more knowings about them \cite{mairesse2007using, 6113107, 6406767, 2018Yuan}. Over the years, it has received much attention; while some studies have applied a combination of linguistic features and machine learning methods \cite{Majumder2017, Sewwandi2017, daSilva2018, YuanCu2018}, some others have focused solely on the machine (deep) learning methods \cite{YuJianguo2017, SunXian2018, yilmazTun2019, ELDASH2021}. In the last few years, we have witnessed a considerable rise in text-based APP, that have used embedding methods to transfer the text elements to a more meaningful space (rather than character space), in favor of better exploitation of computational methods \cite{REN2021102532, xue2018deep, JEREMY2021416, christian2021text}.

Generally it can be inferred that, all of the investigations are intended to acquire more \textit{knowings} about the text elements; each of which through applying miscellaneous methods. Indeed, they are absolutely right; namely, the knowings will be the basis of predictions.

Although various researches have been carried out on text-based APP, no study has been found that essentially is intended to focus on \textit{Knowledge Representation} (\textit{KR}). This paper for the first time (to the best of our knowledge), calls into question the application of knowledge representation and thereby \textit{knowledge graph} in text-based APP. Specifically, this study makes a major contribution to research on automatic personality prediction by proposing a novel knowledge graph-enabled system. Indeed, it meticulously investigates knowledge representation as a novel solution for text-based personality assessment. We believe that, there is knowledge and then there is knowing. Therefore, at first we should discover the world behind the words. It will provide an important opportunity to advance the understanding of text elements. In consequence, practically the significance of our method is that it empowers an APP system to achieve a comprehensive representation of the appeared concepts in the input text which entails the knowledge behind them and models the semantic relations among them, in a more comprehensible manner for the machine, as the basis of its predictions. In fact, the proposed method equips machine with the required knowledge to acquire a better understanding of the entailed concepts in the input text, and accordingly achieve better results.

Knowledge representation is a field of artificial intelligence dedicated to representing information about the world in a form that a computer system can utilize to solve complex tasks \cite{bergman2018knowledge}. In fact, knowledge representation is necessary to understand the nature of intelligence and cognition of concepts so well that computers can be made to exhibit human-like abilities \cite{van2008handbook}. Therefore, in the case of expecting human-like abilities from artificial intelligence, it seems that we ought to represent the knowledge of the world for it. Meanwhile, the knowledge graph actually is the outcome of knowledge representation. It organizes the knowledge of concepts in a graph structure and integrates all existing information about them.

Therefor, aimed to design a knowledge graph-enabled text-based APP system, this paper proposes a three-phase approach, that includes:
\begin{itemize}
	\item\textbf{phase 1: pre-processing} that contains four steps of needed pre-processings, i.e. \textit{tokenization}, \textit{noise removal}, \textit{normalization}, and \textit{named entity recognition} to make the input text ready for main processes in next phase.
	\item\textbf{phase 2: knowledge representation}, the main contribution of this study that comprises three steps, i.e. \textit{graph building}, \textit{graph enriching}, and \textit{graph embedding}. In practice, this phase first attempts to build the corresponding \textit{knowledge graph} for a given text, which is a knowledgeable representation of the input text and then enriches it to cover some neglected pieces of knowledge about concepts. At last, the acquired enriched graph is embedded to a more computationally applicable space, to facilitate the computations in next phase.  
	\item\textbf{phase 3: automatic personality prediction} that aims to predict the personality traits for each input text through a multi-label classification model. To do so, four base deep learning models were proposed that includes Convolutional Neural Network (CNN)-based, simple Recurrent Neural Network (RNN)-based, unidirectional Long Short Term Memory (LSTM)-based and Bidirectional Long Short Term Memory (BiLSTM)-based classifiers.
\end{itemize}

This study aimed to address the following research questions:
\begin{itemize}
	\item[\bfseries RQ.1] How does knowledge graph enabling, influence the performance of a text-based automatic personality prediction system?
	\item[\bfseries RQ.2] What are the performances of popular deep learning models, including CNN, simple RNN, LSTM, and BiLSTM in multi-label classification of knowledge graphs' embedding matrices? Which of them outperforms the others?
	\item[\bfseries RQ.3] Does knowledge graph enabling of an APP system, affects equally the predictions in all five personality traits in Big Five model?	
\end{itemize}

The remaining part of the paper proceeds as follows: \hyperref[sec:APP]{section 2} is concerned with the Big Five personality models. \hyperref[sec:LiteratureReview]{Section 3}, provides an overview of text-based APP systems. The proposed knowledge representation-based APP system, meticulously is demonstrated in \hyperref[sec:Methods]{section 4}. \hyperref[sec:Results]{Section 5} presents the findings of this study, and then \hyperref[sec:Disscussion]{section 6} includes a discussion of the implication of the findings as well as responses to the research questions. Finally \hyperref[sec:Conclusion]{section 7}, namely the conclusion gives a brief summary and critique of the findings.

\section{The Big Five Personality Model}
\label{sec:APP}
So far, various personality trait models have been introduced \cite{Matz2016}. In this study the \textit{Big Five} model (Five Factor Model) \cite{BigFive1992} as the most widely accepted trait model that capably correlates with human traits that presented in written language \cite{Moreno2021}, is used. It basically demonstrates the individuals' personality in five categories: \textit{openness}, \textit{conscientiousness}, \textit{extroversion}, \textit{agreeableness}, and \textit{neuroticism}. \textit{OCEAN} is the acronym of the five categories which we shall refer to, also. Each of the five personality trait represents a range between two extremes \cite{2019introductionToPsych}; i.e. extroversion represents a continuum, between extreme extroversion and extreme introversion. To make it more clear, indicating some facets of each traits which occurs in those people with high scores for each trait, may be useful (for more details please refer to \cite{ramezani2021automatic, feizi2021state}):
\begin{itemize}
	\item \textbf{Openness (O)}: an inclination to embrace new ideas, arts, feelings, and behaviors; unconventional; focused on tackling new challenge; wide range of interests and so imaginative.
	\item \textbf{Conscientiousness (C)}: an inclination to be so self-disciplined, well organized and dutiful; careful and hard-working; reliable, resourceful and on time.
	\item \textbf{Extroversion (E)}: an inclination to be outgoing, energetic, assertive and talkative; affectionate, sociable and articulate; enjoys being the center of attention.
	\item \textbf{Agreeableness (A)}: an inclination to agree and accompany the others; altruist and unselfish; friendly, loyal and patient; modest, considerate and cheerful.
	\item \textbf{Neuroticism (N)}: an inclination to experience negative emotions like anxiety, anger, depression, sadness and envy; impulsive and moody; lack of confidence.
\end{itemize}

Furthermore, it is worth noting that the Big Five traits are mostly independent \cite{2019introductionToPsych}. It means that, being cognizant of someone's one personality trait, does not provide so much information on the remaining traits of the Big Five model.

\section{Literature Review}
\label{sec:LiteratureReview}
In recent years, there has been an increasing amount of literature on APP, that mainly pays particular attention to predict the personality from \textit{text}, \textit{speech}, \textit{image}, \textit{video}, and \textit{social media activities} (likes, visits, mentions, digital footprints, profile interpretation and etc.).

\textit{Text} as an appearance of human language, would competently reflect the writer's personality \cite{MORENO2021110818}. Due to the fact, it is always a matter of concern for personality psychologists. Spreading the Internet-based communication infrastructures, increased the text-based communications among people. It opens the door for computational psychologists to investigate the personality of writers from exchanged texts. Here, we will review the researches conducted on text-based APP.

Taking a glimpse into the reported investigations in APP, it can be claimed that generally all of them are intended to acquire more meaningful and knowing-full alternatives to input text's elements (namely words, terms or generally all the appeared concepts), to deal with. In simple words, dealing with more meaningful alternatives that convey more knowings and information rather than pure character strings, is highly preferred. Actually, these knowings about written language elements, may better represent the knowledge behind them and may lead better predictions about writer's personality. Tracing the evolution of text-based APP systems, sheds more light on this claim. With respect to this claim, generally we can classify the previous studies into five categories: \textit{lexicon-based methods}, \textit{hybrid methods} (combination of lexicon-based and deep learning-based methods), \textit{embedding methods}, \textit{ensemble modeling methods}, and \textit{network-based methods}. A detailed analysis of these categories, are given bellow.

$\blacksquare$ \textbf{Lexicon-based methods} \cite{mairesse2007using, 2018Yuan, 6113107, 6406767, HAN2020105550}: rudimentary techniques, have mainly tended to utilize lexicons which provide linguistic and statistical knowings about text elements. Lexicon-based methods, primarily try to predict the personality of writer through assigning his/her words to pre-determined categories. Linguistic Inquiry and Word Count (LIWC) \cite{pennebaker2001liwc} is one of the most common tools that counts words in psychologically meaningful categories and calculates the degree to which people use different categories of words. It is simply a dictionary of words and word stems, each of them belonging to a one or more category. Given a text, LIWC calculates the percentage of included words in each category. The main idea behind LIWC is that, the word usage in everyday language, reveals the thoughts, personality and feeling of individuals. There have been different versions available, since 2001\footnote{Further information are available at https://liwc.wpengine.com/} and there are more than 80 categories in LIWC2015. Mairesse features \cite{mairesse2007using} and Structured Programming for Linguistic Cue Extraction (SPLICE) are another options that provides linguistic features for words.

In their analysis of APP from the words that people use, Yuan et al. \cite{2018Yuan} have investigated the personality of the characters in vernacular novels. They have created a vector for each dialog using LIWC features which reflects the psychology of the characters. Finally, the vectors have been mapped to the Big Five personality traits, to predict the final personality labels. Mairesse et al. \cite{mairesse2007using} have also investigated a miscellaneous variety of lexicon-based features in order to predict the Big Five personality traits from written text and spoken conversation. 

Among the first reports on APP from the social media text, Golbeck et al. \cite{6113107} have considered LIWC features over the 167 samples of Facebook text contents as well as the users' profile information. The results confirmed limited improvements in APP. In a same manner, the authors in \cite{6406767} have studied the Big Five personality traits from Twitter posts besides the users' profile attributes. They have actually intended to find anti-social traits of narcissism, Machiavellians and psychopathy, (commonly referred to as the dark triad) through using LIWC features. Perusing the reported results, implies that prediction of personality traits from social media text, using lexicon-based methods could not considerably improve the APP accuracy.

Later, in a study which has set out to predict personality traits from social networks' microblogs, Han et al. \cite{HAN2020105550} have found that the context-based knowledge of words may be advantageous to personality prediction. They have believed that since the traditional psychological lexicons (like LIWC) are appropriate for formal texts, they could not efficiently applied in social networks' informal texts. Therefore they have proposed an approach to automatically extract a personality lexicon from social networks, through using keyword extraction techniques and then semantically clustering the extracted keywords. At last, they have simply combined the extracted lexicon (as a prior-knowledge source) with the word embedding vectors and have fed them into a classification model, to predict the Big Five's personality traits' labels. They have partially enhanced the prediction accuracy, even though they have just took the advantage of words' lexical knowledge.

$\blacksquare$ \textbf{Hybrid methods} \cite{Majumder2017, YuanCu2018, SunXian2018, xue2021semantic}: generally in the literature, there seems to be no a tendency among researchers to use lexicon-based methods, solely. Telling the truth, it is hardly fair to shift all the APP responsibility solely to them, because of their superficial knowledge of text elements. Consequently, a large and growing body of literature has investigated the combination of lexicon-based methods with more knowing-full methods that fairly has improved proportionally the predictions' accuracy.

Designing a convolution neural network (CNN) which uses the document-level Mairesse features (extracted from the input text) in an inner layer, have formed the central focus of a study by Majumder et al. \cite{Majumder2017}. They have trained a separate identical binary classifier for each five personality trait in Big Five model that receive sentences of the input text one by one and then aggregate them into a document level vector. Besides, they have finally ignored all the emotionally neutral sentences, to improve the performance. Yuan et al.  \cite{YuanCu2018} have carried out a study to predict the personality of users from their Facebook status contents. Actually, they have combined the LIWC features with deeper features that have been extracted through a deep learning model. They firstly, have extracted the language features via the LIWC tool, and then using a CNN they have automatically extracted the features from textual contents. Subsequently, the two extracted features have been combined to predict the personality labels.

In an another investigation into APP from texts in online social networks, the authors in \cite{SunXian2018} have proposed a bidirectional LSTM model, called 2CLSTM. In order to detect user's personality using the structures of texts, the model has been strengthened by a CNN as well as a latent sentence grouping module which has been applied to capture closely connected sentences. Xue et al. \cite{xue2021semantic} studied the effects of semantic representation of words in APP systems. They acquired a word-level semantic representation of text elements and then fed them into a neural network to obtain higher-level semantics of text elements.

$\blacksquare$ \textbf{Embedding methods} \cite{REN2021102532, xue2018deep, christian2021text, MehtaYash2020, JEREMY2021416, ELDEMERDASH2020, Jiang2020}: alongside these researches, many attempts have been made with the purpose of utilizing complex methods that use even more knowing-full alternatives for text elements. Indeed, they have succeeded in achieving better results during predictions. They mostly pay particular attention to embedding methods, which transform the text elements from a textual space to a real valued vector space. Overall, these methods despite their variety, have better performance in APP, rather than previously mentioned methods. This ability is a consequence of embedding methods' adroitness in meaning acquisition and representation. In a study which has set out to detect personality based on text content analysis, Ren et al. \cite{REN2021102532} have investigated a novel multi-label personality prediction learning model which combines emotional and semantic features. In particular, they have leveraged a Bidirectional Encoder Representation from Transformers (BERT), to generate sentence-level embeddings for extracting semantic features from text, as well as a sentiment dictionary for the sake of text sentiment analysis purposes. Encoders, primarily are designed for achieving a knowing-full representation of input text. They have used the Myers-Briggs Type Indicator (MBTI) and Big Five personality trait models in their study. Xue et al. \cite{xue2018deep}, have also designed a deep learning-based method for personality prediction from text which are posted in online social networks. They have recommended AttRCNN, a hierarchical model that uses a sentence level encoder that is followed by a document level encoder in order to achieve the deep semantic features of text posts. Moreover, they have concatenated the deep semantic features with the statistical linguistic features obtained directly from the text posts, and have fed them into a regression model to predict the Big Five personality traits' labels. Exploiting the embedding methods abilities, in their study Christian et al. \cite{christian2021text} have suggested a multi model deep learning architecture for personality prediction which was combined with various pre-trained language model including BERT, RoBERTa, and XLNet as a feature extraction method on social media text. The main idea behind their investigations was that, since the common deep learning models such as recurrent neural networks (RNNs) and LSTMs suffer from some drawbacks that are defeated using embedding methods, the embedding methods practically outperform them. Specifically, they mostly suffer long training times and inability to capture the context-based information of words and thereby the true meaning of words. At last, the final predictions have taken based on averaging the output of different pre-trained models. Other researches (\cite{MehtaYash2020, JEREMY2021416, ELDEMERDASH2020, Jiang2020}) have also investigated designing embedding-based APP models that make predictions from text.

$\blacksquare$ \textbf{Ensemble modeling methods} \cite{ramezani2021automatic, ELDASH2021, Kunte2020, Ashima2019ensembel, kazameini2020}: meanwhile, taking the advantages of several classifiers and benefit their prediction abilities simultaneously, was a matter of concerns for some studies. Utilizing different APP models predictions, the authors in \cite{ramezani2021automatic} have proposed an ensemble modeling method. Specifically, they have suggested five separate APP models, including term frequency vector-based, ontology-based, enriched ontology-based, latent semantic analysis-based, and deep learning-based (BiLSTM) methods. Then, all of the individual five models have been ensembled through a Hierarchical Attention Network (HAN) as the meta-model. In consequence, they have benefited the ability of five distinct APP model, to make the final decisions about Big Five personality traits. In their study, El-Demerdash et al. \cite{ELDASH2021} have suggested a transfer learning-based APP method that have got the benefits of leading pre-trained language models such as Elmo, ULMFiT, and BERT. To raise the overall personality prediction performance, they have applied a model consists of fusion strategies on data level and classifier level. Adopting the tree pre-trained models, they have used the fusion of Essays and myPersonality datasets for further fine-tuning of the proposed models. Using independent classifiers, each model performs APP, separately. Then, the results have fed into an ensemble learning model that combines multiple classifiers' outputs, to acquire more reliable prediction. Having the same objectives, other researchers (\cite{Kunte2020, Ashima2019ensembel, kazameini2020}) have questioned the usefulness of such an approach.

$\blacksquare$ \textbf{Network-based methods} \cite{sun2019group, 9172426}: there are also a number of investigations that have aimed for achieving a different representation. They mainly have focused on modeling the network among the online social media users. The first report on group-level personality prediction was conducted by Sun et al. \cite{sun2019group}. They have proposed an unsupervised feature learning method called AdaWalk that takes the advantage of independence from labeled dataset. Actually, it was designed based on Network Representation Learning (NRL) method, which was suggested by the authors. Practically, it constructs a complete graph in which its vertices are the users. The graph also posses the generated texts for each user, the similarity between each users' texts, as well as the personality labels in Big Five model. Subsequently, applying random walks (AdaWalks) on the graph, they have transformed the network to a set of sequences and finally have predicted their personality labels after embedding all of them. In the same vein, Guan et al. \cite{9172426} have suggested personality2vec which predicts the personality labels based on NRL using online social networks' texts. The authors have intended to fully utilize the semantic, personality based, and structural information of user generated texts.

Regarding the evolution, the aforementioned claim that all of the contributions have been attempted to achieve more meaningful alternatives for text elements to deal with, would thus seem to be defensible. Actually, the contributions provide strong experimental evidences that more knowing-full alternatives for text elements, may lead more reliable results. What is not yet clear is the impact of an approach which is fundamentally based on knowledge representation of text elements, on APP. An approach which provides really knowledgeable alternatives for text elements that conveys all of the related information and knowings about the concepts as well as their relations.

\section{Material and Methods}
\label{sec:Methods}
The purpose of knowledge representation approach is to demonstrate the cognitive perceptions behind the key concepts in the world, as well as the relations among them. The dexterity of intelligent functionality, is remarkably correlated with existed represented knowledge, both for human and seemingly for machine. We thus primarily decided to represent the knowledge behind the input text elements in favor of APP objectives. To do so, it was decided to manipulate RDF modeling. The aforementioned abilities of RDF model, justify its competency in knowledge representation.

\subsection{Dataset and Some Statistics About It}
In this study, the provided essays in Essays Dataset \cite{pennebaker1999linguistic} were used for training and testing the proposed APP model. It consist of 2,467 essays which are written by psychology students. Afterwards, they were asked to fill out the Big Five Inventory Questionnaire. At the end for each essay, a binary label was assigned to each five personality traits. Throughout this paper, each individual essay will be referred to as \textit{text}. Moreover, it should also be noted that the Big Five personality model was used all over the investigations.

\begin{figure}
	\centering
	\includegraphics[width=0.7\textwidth]{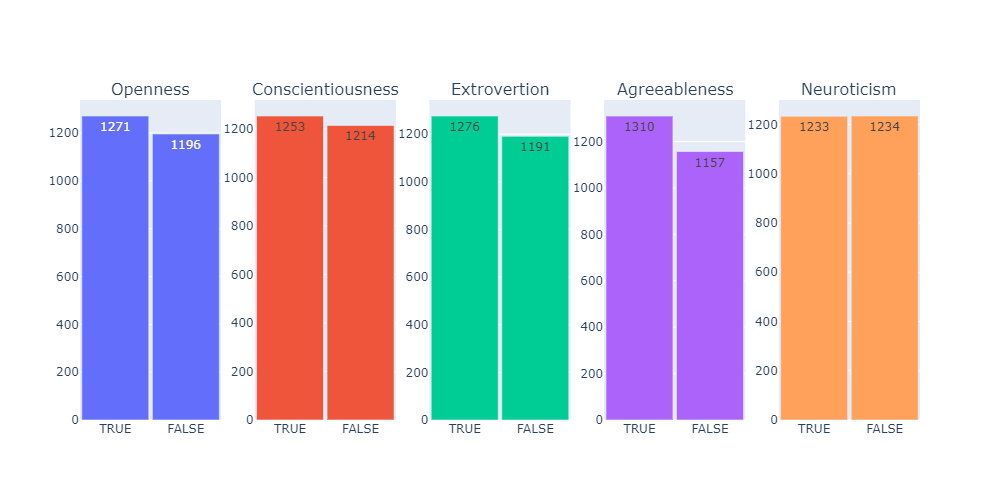} 
	\caption{The distribution of labels in each five personality traits in Essays Dataset}
	\label{fig_LabelDistributionInEssaysDataset}
\end{figure}

Let to scrutinize much more information about the Essays Dataset. \hyperref[fig_LabelDistributionInEssaysDataset]{Figure~\ref*{fig_LabelDistributionInEssaysDataset}}, depicts the distribution of True and False labels throughout the dataset individually in each of five personality traits. Slight difference between the number of True and False labeled essays, reveals that the dataset is balanced and appropriate for learning the APP model.

\begin{figure}
	\centering
	\includegraphics[width=0.5\textwidth]{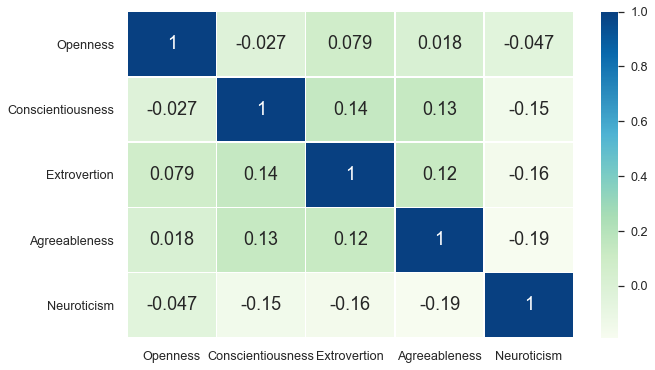}
	\caption{The correlation matrix for five personality traits in Essays Dataset}
	\label{fig_EssaysDatasetCorrelationMatrix}
	
\end{figure}

\hyperref[fig_EssaysDatasetCorrelationMatrix]{Figure~\ref*{fig_EssaysDatasetCorrelationMatrix}}, compares the correlations among the five personality traits in Essays Dataset. As it can be seen, a correlation matrix is a symmetric matrix which all the values on the main diagonal are equal to 1. The correlation coefficient can range between -1 and +1. The larger the absolute value of coefficient, implies stronger relationship between two traits. Specifically, a positive coefficient between two traits means that, being aware of one trait's label, allows a correct prediction of the other; as close as possible to +1, it will conclude more correct predictions.

\begin{figure}
	\includegraphics[width=\textwidth]{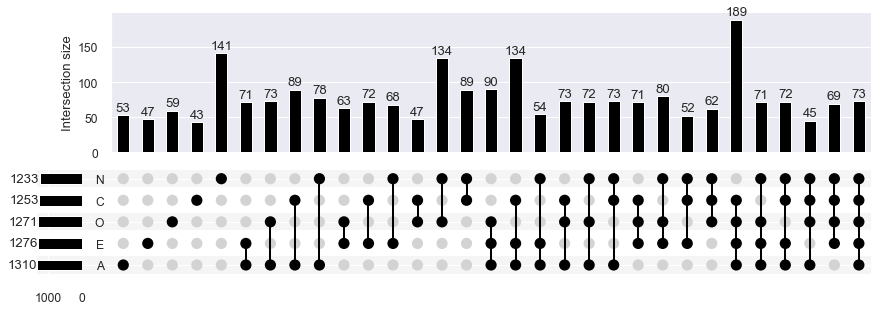}
	\caption{The UpSet plot of intersections among \textit{true labeled sets} of personality traits in Essays Dataset \newline\scriptsize notes: \textit{i}$)$ sets \{O, C, E, A, N\} are sorted by their cardinality in ascending order, \textit{ii}$)$ light (empty) circles indicate that the set is not part of that intersection}
	\label{fig_Upset_Chart}
\end{figure}

The UpSet \cite{6876017} plot of five sets of personality traits is presented in \hyperref[fig_Upset_Chart]{Figure~\ref*{fig_Upset_Chart}}. An UpSet plot actually is considered as a substitution for Venn diagram, when dealing with more than 3 sets. Having five sets of personality traits (namely \textit{O}, \textit{C}, \textit{E}, \textit{A}, and \textit{N}), the UpSet plot make it possible to provide an efficient way to visualize the intersections of five sets. Each row at the bottom of the \hyperref[fig_Upset_Chart]{Figure~\ref*{fig_Upset_Chart}} denotes to a set and each column corresponds to one segment in Venn diagram, depicted with five light or black circles. A black circle indicates that the corresponding set is participating in the intersection, and a light circle vice versa. Indeed, a light circle indicates that the complement of the set $(\acute{O}, \acute{C}, \acute{E}, \acute{A}, or \acute{N})$ is participating in the intersection. In particular, the rightmost column which has five black circle for all of the five sets, is equal to (\textit{O$\cap$C$\cap$E$\cap$A$\cap$N}). The bar chart on top of the \hyperref[fig_Upset_Chart]{Figure~\ref*{fig_Upset_Chart}} represents the cardinality of each corresponding intersection. It worth mentioning that the plot, depicts the intersections among true labeled essays' sets. That is to say, just the true labeled essays in \textit{OCEAN} traits are taken into consideration.

\subsection{System Architecture}
\begin{figure}
	\includegraphics[width=\textwidth]{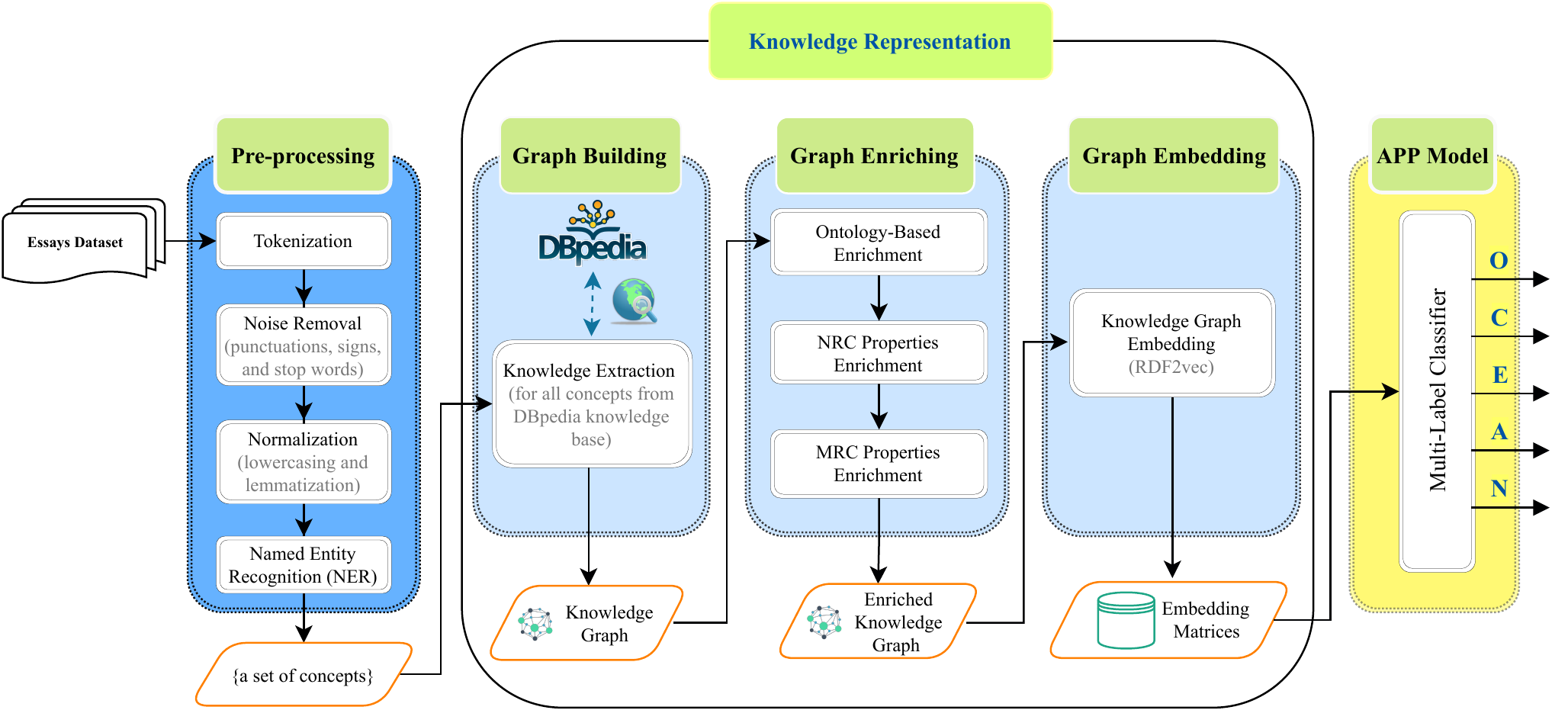}
	\centering
	\caption{System Architecture}
	\label{fig_System_Architecture}
\end{figure}

Aimed to answer the research questions stated at the beginning of this study, we suggested a three-phase approach which is outlined in \hyperref[fig_System_Architecture]{Figure~\ref*{fig_System_Architecture}}. The experiment proceeds following the phases bellow.

\subsubsection{Phase 1: Pre-processing}\mbox{}\\
In this phase the aim was to clean and transform the input texts into a more digestible form for machine to be processed in next phase. This traditional prominent and common practice in natural language processing, basically consists of miscellaneous activities depending on the existing task. What follows is a description of pre-processing activities were carried out in first phase, as depicted in \hyperref[fig_System_Architecture]{Figure~\ref*{fig_System_Architecture}}.

\paragraph{$\bullet$ Tokenization:}
having a text, ``\textit{tokenization}'' is the task of chopping it up into pieces called \textit{tokens} which roughly correspond to words \cite{schutze2008introduction}. Tokens are also deemed as the smallest useful semantic unit for processing. For this purpose, the tokenizer which is provided by Natural Language Toolkit (NLTK) \cite{bird2009natural}, was used.

\paragraph{$\bullet$ Noise Removal:}
in the interest of achieving more plain text, it was necessary to remove undesirable and interfering pieces of input text. Regarding the current task, we removed punctuations, signs and stop words using NLTK.

\paragraph{$\bullet$ Normalization:}
``\textit{normalization}'' is the process of canonicalizing tokens to a more uniform sequence so that matches occur despite superficial differences in the character sequences of the tokens \cite{schutze2008introduction}. It practically decreases the amount of information that the machine has to deal with; those that conceptually are similar, but morphologically different. In an attempt to normalize the input text, \textit{lowercasing} and \textit{lemmatization} were carried out.

\textit{Lemmatization} is the morphological analysis of words that groups together their inflected forms and returns the bases or dictionary forms of them, which are called \textit{lemma}. Since lemmatization converts words to their meaningful dictionary form and yields the correct form of concepts which really exist in the world\footnote{compared to \textit{stemming} which is an alternative method to reduce inflected words to their \textit{stems} and usually is fulfilled through chopping off the ending characters of the word; it usually returns incorrect and misspelled forms of words.}, it is appropriate in current task. The resulted meaningful concepts will be queried during knowledge graph building in the following phase. In this study, the lemmatization was also carried out using NLTK.

\paragraph{$\bullet$ Named Entity Recognition (NER):}
with an eye to achieving the knowledge behind the words, it will be necessary to recognize the \textit{named entities} from the input text. The sequences of words which actually are the name of things (that is to say, the name of organization, person, company, event and etc.). As a matter of fact they convey more information than do the other words. In current study, the spaCy NER \cite{spacy} was used to recognize named entities.

After the completion of pre-processing phase, what is extant constitutes a set of concepts which convey the fundamental notions that have appeared in the input text. It should be mentioned that after the NER, duplicate elements were removed from the set of concepts. Then, in order to prepare the elements to be matched with DBpedia knowledge base entries, first letter capitalization was performed for all of the elements, and white space replacement with underscore was done for multi-word elements, as well. Now, everything is ready to find out the world behind the words.

A brief summary of the phase 1, as it can be seen in \hyperref[fig_System_Architecture]{Figure~\ref*{fig_System_Architecture}}, may be described as follows:
\begin{enumerate}[i)]
	\setlength{\itemsep}{0pt}
	\setlength{\parskip}{0pt}
	\item input: essays' texts from Essays Dataset;
	\item output: a set of extracted concepts for each text;
	\item objective: to prepare a more digestible form of input text for main processes in next phases.
\end{enumerate}

\subsubsection{Phase 2: Knowledge Representation}\mbox{}\\
\label{sec:phase2KR}
As it was stated in \hyperref[section-1]{Introduction}, we chose the graph structure to represent the existing knowledge of concepts in the input text, as well as the relations among them, and to eventuate a knowledge graph for each text. The two first steps in bellow, fully describe that how a knowledge graph was built for each text. The purpose of the current phase is to attain a comprehensive representation of existing knowledge of the input set of concepts, so that it could be applied for subsequent computations. Hence, the resulted knowledge graph was transferred to a numerical space using a graph embedding method in third step. The suggested three-step procedure, is shown in \hyperref[fig_System_Architecture]{Figure~\ref*{fig_System_Architecture}} and proceeds as follows.

\paragraph{Step 1: Knowledge Graph Building}\mbox{}\\
As a matter of fact, the set of extracted concepts from the input text in phase 1, substantially organizes the existent notions in it. There is always knowledge behind every concept. The current step is intended to extract the knowledge of appeared concepts in input text from \textit{DBpedia} knowledge base \cite{DBpedia2007} and then tries to establish a \textit{knowledge graph} which effectively organizes and represents the knowledge of containing text elements \cite{CHEN2020KG}. 

A knowledge graph is a large-scale knowledge base composed of a large number of entities (objects, events, or concepts) and relationships between them \cite{ChenZhe2020}. Actually, it is a directed heterogeneous (having vertices/edges of different types) labeled multi-graph (a graph which is allowed to have multiple directed edges between the same pair of vertices), in which the labels have well-defined meanings \cite{StanfordKG2021}. The graph structure in knowledge graph, adroitly posses what is needed in knowledge representation \cite{chein2008graph}. Like all graphs, it consists of vertices and edges which the vertices represent the entities of real world, and the edges connects pairs of vertices according to their relationship. What is more, the labels convey exact information (sometimes called \textit{semantics}) about the existing relationship (edge) between the vertices. The encompassed knowledge in the knowledge graphs stores in the form of \textit{triples} same as (\textit{h, r, t}) that stands for (\textit{head entity}, \textit{relationship}, \textit{tail entity}). That is to say, having a set of vertices \textit{V}, along with a set of labels \textit{L}, the knowledge graph would be a subset of the cross product \textit{V} × \textit{L} × \textit{V}; each member of this set is referred to as a triple \cite{StanfordKG2021}. Each triple may also be interpreted as (\textit{subject}, \textit{predicate}, \textit{object}); for instance (Louvre, is located, Paris). \cite{KnowledgeGraphsBook2021} provides detailed information about knowledge graphs.

Meanwhile, there is a well suited framework that matches as close as possible to the knowledge graph's triple requirement, namely \textit{Resource Description Framework} or \textit{RDF}. In essence, it is a standard for representing information in the Web. Equally, this framework is made up of (subject, predicate, object) triples. A set of RDF triples, that construct an RDF dataset can be also viewed as a directed heterogeneous labeled multi-graph (like a knowledge graph) which is also referred to as \textit{RDF graph} \cite{RDFHiba2018}. In an RDF graph, vertices (subjects and objects) are either IRIs (Internationalized Resource Identifier) which stands for a unicode string representing resources, or \textit{literals} that contain values such as strings, numbers, and dates. As well, the edges (predicates or labels) are also IRIs representing predicates or relationships\footnote{More detailed information about RDFs are available at https://www.w3.org/TR/rdf11-concepts/}.

Intended to build the knowledge graph of input text during \textit{graph building} step in phase 2, as previousely mentioned the DBpedia knowledge base is used. DBpedia actually is a community effort to extract structured information from Wikipedia and to make them available on the Web. The 2016-04 release of the DBpedia, contains 9.5 billion RDF triples, that describe about 6 million entities.

One can easily query on DBpedia dataset online via SPARQL endpoint which is a standard query language and protocol for linked open data and RDF databases. Thus, we queried all the elements of input concepts' set on DBpedia and extracted all the relevant knowledge of each concept. It was fulfilled through ``DESCRIBE'' in SPARQL query language (with no binding in SELECT clause and no pattern in WHERE). Specifically, it asks for a description about queried concept (some times called resource) and receives any concepts or resources which are directly related to the queried concept (for further details about SPARQL query language, please refer to \cite{Hogan2020}). As previously mentioned in section \ref{section-1} the results returned from the queries are in form of RDF triples that provide a set which is also called \textit{RDF graph}. RDF graph, organizes the knowledge of concepts in a directed heterogeneous labeled multi-graph which is wildly known as \textit{knowledge graph}. It almost encompasses all the (existent) knowledge of concepts. One can find the results for a given query \textit{X} on DBpedia at \textit{https://dbpedia.org/page/X}\footnote{The abundance of resulted RDFs for one query, prevented us from exhibiting the concluding results for a sample query. Please note that first letter capitalization and white space replacement with underscore for multi-word concepts, are necessary.}.

\paragraph{Step 2: Knowledge Graph Enriching}\mbox{}\\
After building the knowledge graph for the input text, different pieces of information (likewise in form of RDF triples) enrich the current knowledge graph during this step. Enriching the representation, inevitably gives more focus to some neglected aspects of facts about entities. In other words, having limited aspects of knowledge, will bound the intelligent agent's perception of the world \cite{jakus2013book}. Consequently, the following graph enrichments were carried out on the resulted knowledge graph:

\paragraph{$\bullet$ Ontology-based enrichment:}
\textit{ontology} actually is a branch of metaphysics dealing with the nature and relations of beings \cite{mw_ontology}. It demonstrates that how the things are related to each others in a systematic hierarchical classification.

A great deal of attention must be paid that knowledge bases essentially are made up of \textit{instances}, rather than \textit{concepts}; what are the foundation of ontologies. Therefore, it would indisputably enhance the representation by means of providing a different aspect of knowledge about things.

To do so, we used the DBpedia ontology. It covers 768 classes (a complete list of covered classes is available in \cite{DBpediaOntologyClasses}) which are described by 3,000 properties for about 4,233,000 instances. One can easily find the ontology-based representation of a given concept \textit{X} in DBpedia ontology at \textit{https://dbpedia.org/ontology/X}. At the beginning of foundation, it had been created based on most commonly used infoboxes within Wikipedia (in 2008) before it evolved into a crowd-sourcing effort. All the RDFs resulted from matching the concepts with DBpedia ontology, were added to the previously achieved RDF graph.

\paragraph{$\bullet$ NRC Properties enrichment:}
words can be associated with different intensities of an emotion. The NRC Emotion Intensity Lexicon \cite{NRCAFFECTDic} which is provided by National Research Council Canada (NRC), contains real-valued intensity scores for eight basic emotions (namely anger, anticipation, disgust, fear, joy, sadness, surprise, and trust) for about 10,000 entries in English. The lexicon mainly includes more common English words and terms along with those that are more prevalent in social media. The aim of present section is to enrich the representation of input text through enhancing the eight provided emotions' degrees for included words. The emotions' scores for each concept were added to the existed RDF graph in form of literal RDFs for each word, in case of inclusion in NRC.

\paragraph{$\bullet$ MRC Properties enrichment:}
the final knowledge graph enriching process was the enhancement of psycholinguistic properties to the RDF graph. It was carried out through MRC psycholinguistic database \cite{MRCdb}. MRC is a publicly available machine usable dictionary which contains (up to 26) linguistic and psycholinguistic attributes (like syntactic, phonological, orthographic, and semantic features) for 150,837 English words. These properties also were added to the existed RDF graph in form of literal RDFs for each word.

\paragraph{Step 3: Knowledge Graph Embedding}\mbox{}\\
So far, we have achieved the knowledge graph for a given text that basically is made up of RDFs, for both vertices and edges. This step transforms the resulted knowledge graph into a vector space and produces its equivalent embedding matrix. It strives to maximally persevere graph's structure, even though it practically performs dimensionality reduction on it. In this study, the knowledge graphs were embedded according to the method proposed by Ristoski et al. \cite{ristoski2019rdf2vec}. In their seminal contribution, they proposed \textit{RDF2vec}; a tool for creating vector representations of RDF graphs. RDF2vec actually is inspired by the word2vec \cite{mikolov2013WORD2VEC}, which is a well known \textit{word} embedding method (representing words in numeric vector space). RDF2vec almost works similar to word2vec; the major difference is the input sequence. While word2vec receives a set of sentences for training the learning model as the input sequence, RDF2vec uses random walks on the RDF garaph to create sequences of RDF vertices to feed them into the same learning model. As a consequence, similar vertices place close to each others in the final vector sapce and dissimilar ones, do not; like what happens for words after embedding in word2vec. To put it briefly, in this step the corresponding embedding matrix for a given knowledge graph was achieved.

We set the maximum depth for each walk and the maximum number of walks per entity, both equal to 5 in all the random walks which carried out on knowledge graphs. It should be considered that in practice, the two first phases, namely pre-processing and knowledge representation, were iteratively executed for all of the essays in Essays Dataset (please refer to \hyperref[fig_System_Architecture]{Figure~\ref*{fig_System_Architecture}}), and lasted for more than four months\footnote{Experiments were run on a computer with an Intel i7-7700K processor, using 64 GB of ram and running Windows 10.}. As a result of such iteration, a set of embedding matrices resulted from the knowledge graphs embeddings were achieved; in which, the rows of each matrix, are dedicated to existed concepts in corresponding essay, and the columns are dedicated to the embedding dimensions. The number of rows in each matrix is different depending on the existing number of concepts in corresponding essay. Therefore, to fix the number of rows and achieve embedding matrices with same number of rows, we selected the 10,000 most frequent concepts in all final resulted knowledge graphs for Essays Dataset's essays. The larger number of rows leads to sparsity of embedding matrices and smaller number leads to ignore the included concepts. The number of columns (embedding size) which is specified by RDF2vec, by default is equal to 500.

In consequence, a brief summary of the phase 2, as it can be seen in \hyperref[fig_System_Architecture]{Figure~\ref*{fig_System_Architecture}} may be described as follows:
\begin{enumerate}[i)]
	\setlength{\itemsep}{0pt}
	\setlength{\parskip}{0pt}
	\item input: a set of concepts for each text;
	\item output: equivalent embedding matrix for each text;
	\item objective: knowledge representation for each text; specifically building, enriching and embedding the corresponding knowledge graph for each text.
\end{enumerate}

\subsubsection{Phase 3: Automatic Personality Prediction}
\begin{figure}[!htb] 
	\centering
	\begin{tabular}{|c|c|}
		\hline
		\subcaptionbox{CNN model\label{CNNArchitecture_a}}[0.4\linewidth]{\includegraphics[width=0.8\linewidth]{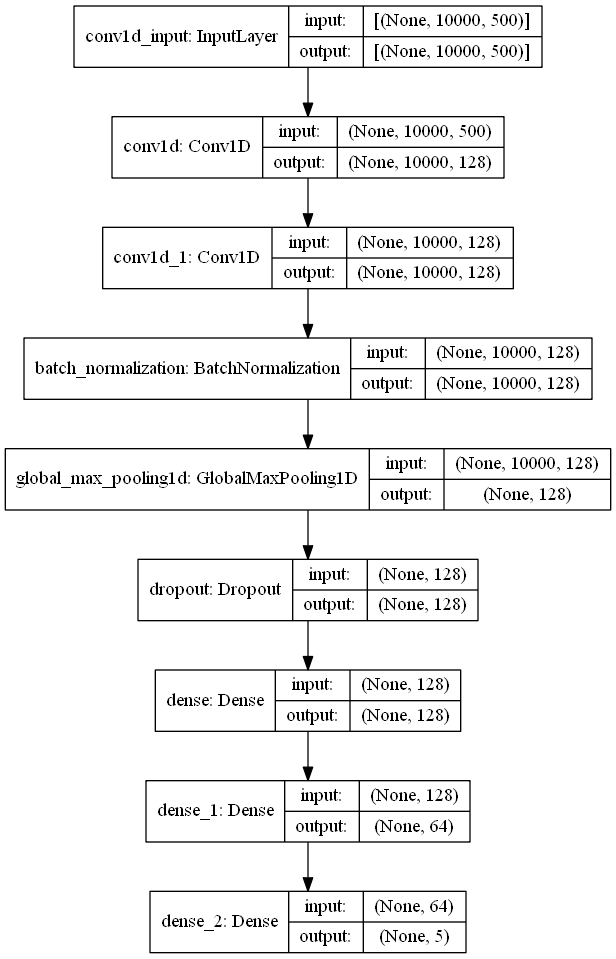}} &
		\subcaptionbox{RNN model\label{RNNArchitecture_b}}[0.4\linewidth]{\includegraphics[width=0.8\linewidth]{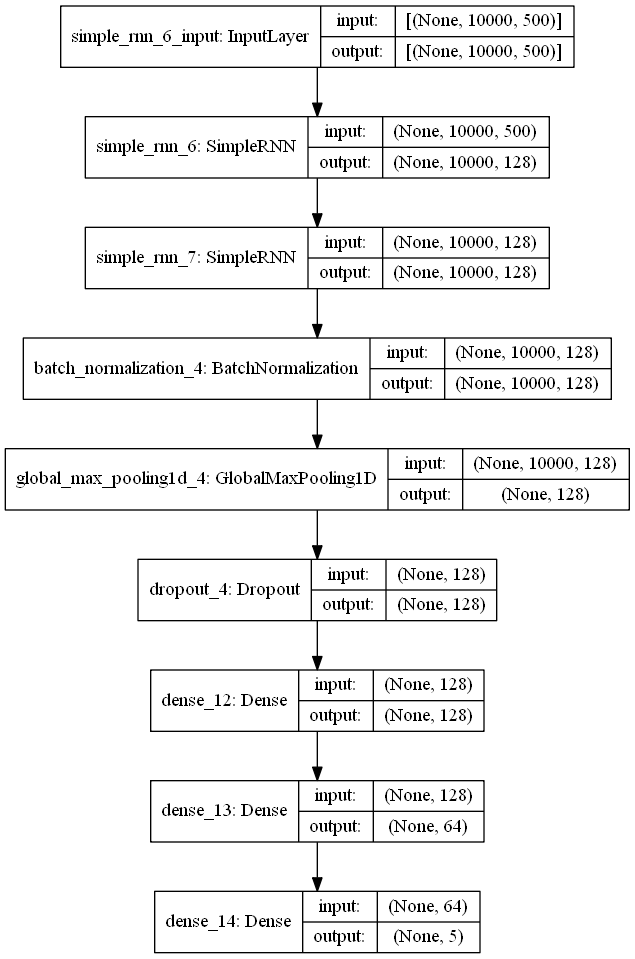}} \\
		\hline
		\subcaptionbox{LSTM model\label{LSTMArchitecture_c}}[0.4\linewidth]{\includegraphics[width=0.8\linewidth]{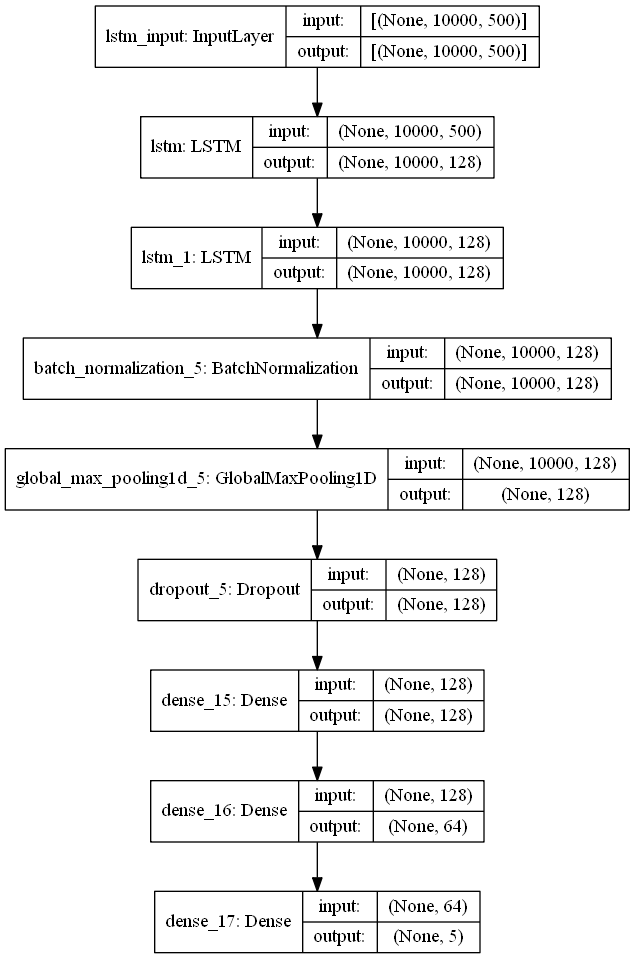}} &
		\subcaptionbox{BiLSTM model\label{BiLSTMArchitecture_d}}[0.4\linewidth]{\includegraphics[width=0.8\linewidth]{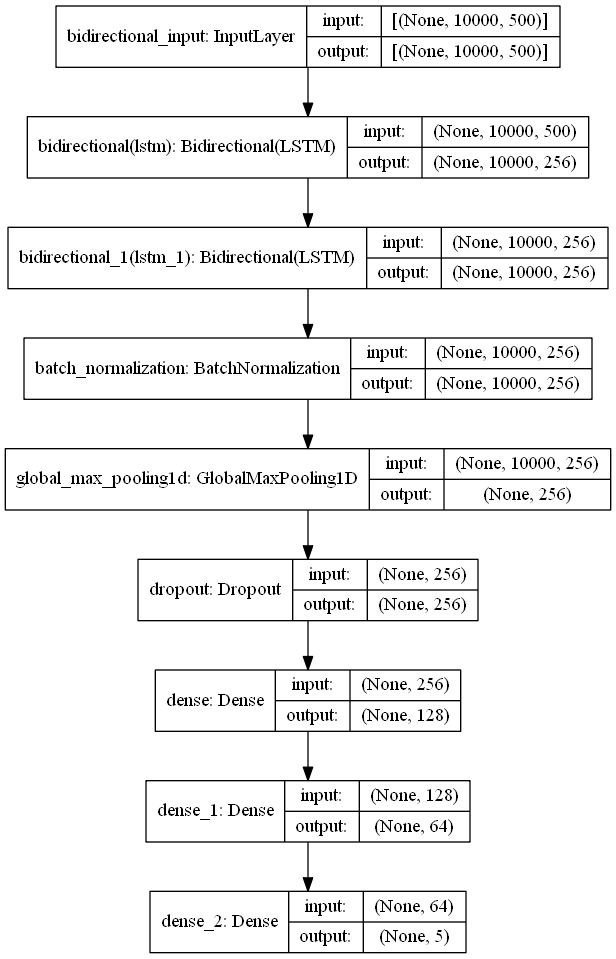}}
		\\
		\hline
	\end{tabular}
	\caption{The Summaries of proposed APP classifiers}
	\label{fig_theArchitectureOfClassifiers}
\end{figure}

\mbox{}\\
Finally, four separate classification models were developed to carry out personality prediction, including: convolutional neural network (CNN)-based, simple recurrent neural network (RNN)-based, long short-term memory (LSTM)-based and bidirectional LSTM (BiLSTM)-based classifiers. To appraise the competency of suggested knowledge graph-enabled APP merely, some of the base and most well known deep learning classification models, with maximally similar architectures and same configurations, were used. Classification in all Big Five traits were fulfilled concurrently. In fact, each model performs a multi-label binary classification which assigns five labels to each of the \textit{OCEAN} traits for a given text. Some common settings which were applied in all suggested APP models, are presented in \hyperref[table:ParameterSettings]{Table~\ref*{table:ParameterSettings}}. Furthermore, As shown in \hyperref[fig_theArchitectureOfClassifiers]{Figure~\ref*{fig_theArchitectureOfClassifiers}} the architecture of each model is composed of two stacked classifiers (like CNN), that leads to better results rather than single classifier. The classifiers are then followed by a batch normalization, to expedite the training and regularize the model. Next, applying a pooling layer as well as a dropout layer, will help to avoid over fitting through providing an abstracted form of the representation. Finally, the models are followed by two consecutive dense layers to classify the extracted features from previous layers, change the dimensions of the vectors, and make possible the final prediction in the output layer.

\begin{table}
	\centering
	\caption{Common parameters' settings among all of the proposed APP models (including: CNN, RNN, LSTM, and BiLSTM)}
	\label{table:ParameterSettings}
	\scriptsize
	
	\begin{adjustbox}{max width=\textwidth}
		
		\begin{tabular}{c c c c||c c c c }
			\toprule
			
			\textbf{Parameter} &  & \textbf{Setting} & & & \textbf{Parameter} &  & \textbf{Setting}  \\
			\cmidrule{1-1}\cmidrule{3-3}\cmidrule{6-6}\cmidrule{8-8}
			Train-Test split ratio (\%) &   & 80-20  & &   & Optimizer &   & Stochastic Gradient Descent (SGD)    \\
			
			Number of epochs &   & 30  & &  & Learning rate &   & 0.01    \\
			
			Early stopping &   & Applied on validation loss  & &  & Loss function &   & Binary\textunderscore crossentropy    \\
			
			
			Patience value &   & 4  & &   & Batch size &   & 32    \\ 
			
			Activation function &   &  Sigmoid  & &  & Cross validation &   & 10-fold   \\ 
			\bottomrule
		\end{tabular}
		
	\end{adjustbox}
	
\end{table}

\paragraph{$\bullet$ Convolutional neural network (CNN)-based classifier}\mbox{}\\
Convolutional neural networks, as a model with impressive performance have been extensively investigated in various problems including visual recognition, speech recognition and natural language processing \cite{CNN2018}. To classify the resulted embedding matrices, a model with two one-dimensional convolutional layers followed by a batch normalization layer, a pooling layer, a dropout layer for regularization, and finally two fully connected layers, was developed. In each convolutional layer, 128 parallel feature maps and a kernel size of 7, along with same padding were applied. \hyperref[CNNArchitecture_a]{Figure~\ref*{CNNArchitecture_a}}, presents a summary of the model.

\paragraph{$\bullet$ Recurrent neural network (RNN)-based classifier}\mbox{}\\
Recurrent neural networks have shown excellent dexterity in text classification tasks. The foundation of RNN \cite{RNN2020}, makes it possible to utilize previous step's outputs as inputs in current step. To rephrase it, while traditional neural networks deal with the inputs independently of one another, RNNs manipulate a set of previous inputs. Furthermore, the internal state of an RNN which acts as memory, empowers it to learn from previous information and grants a privilege of processing the sequential inputs like text. The suggested simple RNN-based classifier, encompasses two simple RNN layers followed by a batch normalization layer, a pooling layer, a dropout layer for regularization, and finally two fully connected layers. A summary of the model is shown in \hyperref[RNNArchitecture_b]{Figure~\ref*{RNNArchitecture_b}}.

\paragraph{$\bullet$ Long short-term memory (LSTM)-based classifier}\mbox{}\\
Long short-term memory networks as a kind of RNNs, are suggested to deal with learning long-term dependencies problem \cite{RNN2020}. In simple terms, simple RNNs suffer one major drawback; they are not able to remember information for a long period of time, what is resolved capably by LSTMs. Two stacked LSTMs followed by a batch normalization layer, a pooling layer, a dropout layer for regularization, and finally two fully connected layers, construct the design of proposed LSTM-based classification model. A plot of model is depicted in \hyperref[LSTMArchitecture_c]{Figure~\ref*{LSTMArchitecture_c}}.

\paragraph{$\bullet$ Bidirectional long short-term memory (BiLSTM)-based classifier}\mbox{}\\
Indeed, as can be inferred BiLSTMs are a bidirectional form of LSTMs. In simple words, LSTM is a unidirectional network which utilizes previous information that have already passed through it in forward direction within sequence processing, while a BiLSTM network exploits both previous and future information in forward and backward directions, respectively. Telling the truth, it consists of two LSTMs: one analysis the input sequence from beginning to the end in forward direction and the other one, from end to beginning in backward direction \cite{BiLSTM2019}. The final output is the concatenation of the two LSTMs. Two stacked BiLSTMs, followed by a batch normalization layer, a pooling layer, a dropout layer, and finally two fully connected layers, comprises the architecture of the proposed BiLSTM-based classifier. \hyperref[BiLSTMArchitecture_d]{Figure~\ref*{BiLSTMArchitecture_d}}, depicts a summary of the model.

At last, a brief outline of phase 3 as it can be seen in \hyperref[fig_System_Architecture]{Figure~\ref*{fig_System_Architecture}} is as follows:
\begin{enumerate}[i)]
	\setlength{\itemsep}{0pt}
	\setlength{\parskip}{0pt}
	\item input: a set of embedding matrices;
	\item output: predicted labels for \textit{OCEAN} traits for each embedding matrices;
	\item objective: personality prediction using multi-label classification model.	
\end{enumerate}

\hyperref[Algorithm:General-Architecture]{Algorithm~\ref*{Algorithm:General-Architecture}}, details a step-by-step flow of the proposed method that would assist towards a better comprehension of the method.


\begin{algorithm}
	\caption{Algorithm of the proposed method}
	\label{Algorithm:General-Architecture}
	\scriptsize
	
	\begin{algorithmic}[1]
		\MyFor{essay $e_{i} \in$ Essays Dataset}
		\State Phase-1: Perform \textit{pre-processing} activities, including:
		\begin{enumerate}[(i)]
			\setlength{\itemindent}{2.5em}
			\setlength{\parskip}{0pt}
			\item Tokenization
			\item Noise removal (punctuations, signs, and stop words)
			\item Normalization (lowercasing and lemmatization)
			\item Named Entity Recognition (NER)
		\end{enumerate}
		\State Phase-2: Perform \textit{knowledge representation} for $e_{i}$, more specifically:
		\begin{enumerate}[(i)]
			\setlength{\itemindent}{2.5em}
			\setlength{\parskip}{0pt}
			\item Build the corresponding knowledge graph (KG) for $e_{i}$, through DBpedia
			\item Enrich the acquired KG using DBpedia ontology, NRC, and MRC
			\item Embed the acquired enriched KG using RDF2vec, and save it in \textit{\textbf{Embeddings}} set
		\end{enumerate}
		\EndMyFor
		\State Using the \textit{\textbf{Embeddings}} set, train and test:
		\begin{enumerate}[(i)]
			\setlength{\itemindent}{2.5em}
			\setlength{\parskip}{0pt}
			\item A CNN-based  multi-label binary classification model to perform APP
			\item An RNN-based  multi-label binary classification model to perform APP
			\item An LSTM-based  multi-label binary classification model to perform APP
			\item A BiLSTM-based  multi-label binary classification model to perform APP
		\end{enumerate}
	\end{algorithmic}
\end{algorithm}

\section{Results}
\label{sec:Results}

\subsection{Evaluation Metrics}
Traditionally, classification models are evaluated through some well known evaluation metrics including \textit{precision}, \textit{recall}, \textit{f-measure}, and \textit{accuracy} \cite{schutze2008introduction}. There are two determining sets which play crucial role in their values, specifically the set of essays' ``actual labels'' which is sometimes referred to as \textit{gold standard} and the set of ``system predicted labels''. Practically, for each prediction in a given class (namely \textit{O}, \textit{C}, \textit{E}, \textit{A}, and \textit{N}) there are four possible combinations of actual labels and system predicted labels, including:
\begin{enumerate}[i)]
	\setlength{\itemsep}{0pt}
	\setlength{\parskip}{0pt}
	\item \textit{TP} (True Positive): that occurs when the actual label is true and the system predicted label is also true;
	\item \textit{TN} (True Negative): that occurs when the actual label is false and the system predicted label is also false;	
	\item \textit{FP} (False Positive): that occurs when the actual label is false while the system predicted label is true;	
	\item \textit{FN} (False Negative): that occurs when the actual label is true while the system predicted label is false.
\end{enumerate}

Essentially, in an APP system evaluation the TP and TN play a leading part, due to the fact that in such classification systems it is prominent to truly predict that a given text really belongs or not belongs to the class. In both of TP and TN the system predicted labels are equal to the actual labels, hence it can be stated that the total number of TPs and TNs denotes to the APP system's correct predictions. Consequently, the ratio of systems correct predictions to the total number of predictions, reveals the quality of prediction, what it actually is known as \textit{accuracy}. That is to say, $accuracy = (TP + TN)/ (TP + TN + FP + FN)$. 

Moreover, the precision and recall as well as their weighted harmonic mean which is called f-measure convey some facts about the performance of classification system. \textit{Precision} (\textit{P}), mainly concerns system's true labeled predictions. It reveals that, what proportion of system's true labeled predictions, has actual true labels. In other words, $P = TP/(TP + FP)$. This is while, \textit{recall} (\textit{R}) mainly concerns true labels in gold standard. It tries to reveal that what proportion of true labeled samples in gold standard has achieved true labels, after the system prediction. It means that $R = TP/(TP + FN)$.

Both of the precision and recall are unreliable metrics in classification systems' evaluation when they are considered separately. To put it another way, there may be some cases with high values of precision and low values of recall simultaneously, and vice versa. It is principally because of their partial coverage and incomplete reports. Hence, f-measure is suggested to address this problem. In fact, it makes a trade off between precision and recall and combines their included facts; precisely, $f-measure = (2\times P\times R)/(P + R)$. However, it still suffers from a significant drawback. Actually, TN as a prominent factor in evaluation is completely neglected. As an illustration, it ignores all of the correctly false labeled samples by system. Thereby, accuracy is preferred to f-measure in APP system evaluation.

\subsection{Evaluation Results}
\begin{figure}[htb]
	\centering
	\begin{tabular}{cc}
		\subcaptionbox{Knowledge graph's vertices (provided by Gephi, the ForceAtlas2 algorithm).\label{fig_sample_KG_a}}[0.5\linewidth]{\includegraphics[width=0.8\linewidth]{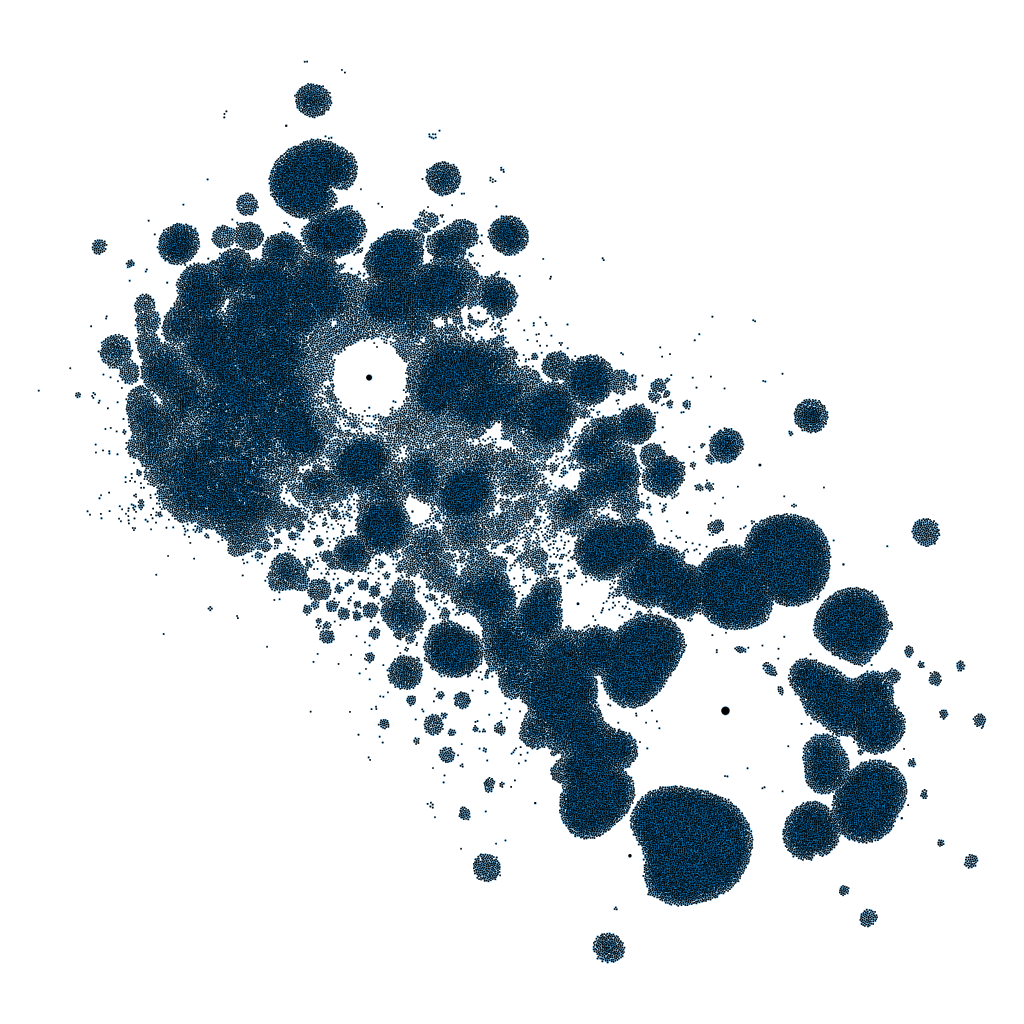}} &
		\subcaptionbox{Embedded knowledge graph in 2D spcae.\label{fig_sample_KG_b}}[0.5\linewidth]{\includegraphics[width=0.8\linewidth]{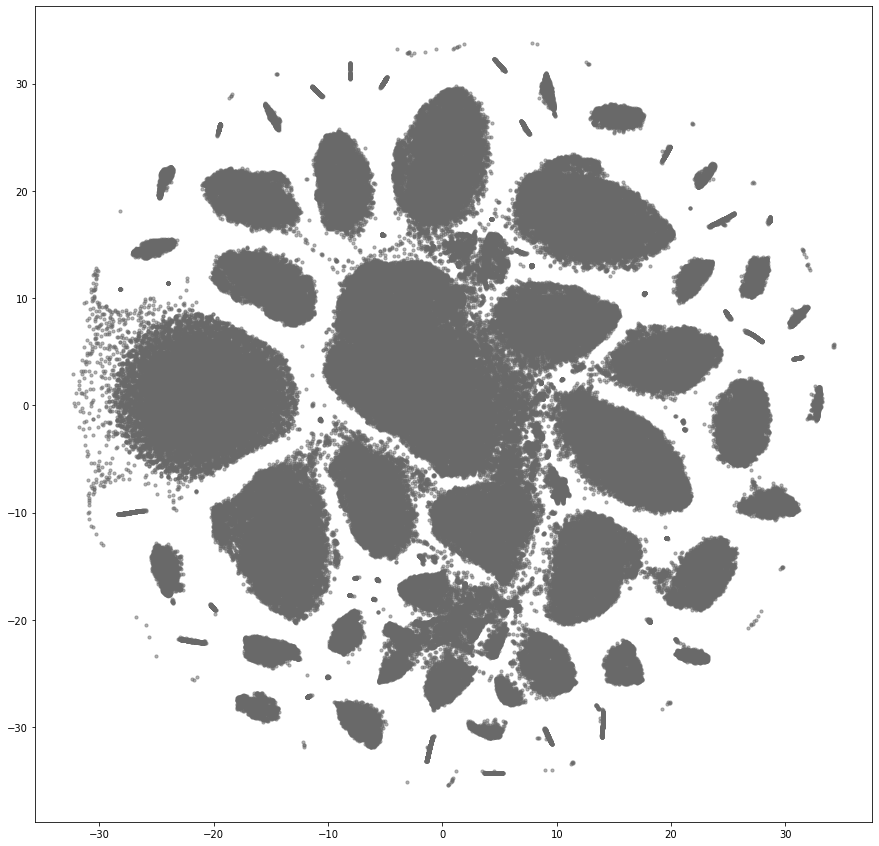}}
	\end{tabular}
	\caption{The resulted final knowledge graph from phase 2 for a sample essay (2004\textunderscore139) in Essays Dataset (including 226,763 vertices and 532,146 edges). The edges and labels in knowledge graph are emitted for better visualization.}
	\label{fig_phase2_OutputsForaSampleEssay}
\end{figure}

This study was undertaken to design a knowledge graph-enabled automatic personality prediction system, and evaluate the efficacy of knowledge graph-enabling of a personality prediction system. Accordingly, a three-phase approach was proposed which by receiving a text proceeds to carry out some pre-processings in first phase, and then build, enrich, and embed the corresponding knowledge graph consecutively in second phase, as it completely scrutinized in \hyperref[sec:phase2KR]{section~\ref*{sec:phase2KR}}. \hyperref[fig_phase2_OutputsForaSampleEssay]{Figure~\ref*{fig_phase2_OutputsForaSampleEssay}}, provides the results obtained from the second phase for a sample essay in Essays Dataset. Eventually, the resulted embedding matrix was classified through four independent classification models in third phase and the predicted labels in each \textit{OCEAN} traits were assigned. This section summarises the findings and contributions made.

Specifically, there were suggested four APP classification models, namely CNN-based, RNN-based, LSTM-based and BiLSTM-based classifiers. However that accuracy outperforms precision, recall and f-measure in APP systems' evaluation, we will report the evaluation results for all of them. Albeit, we will mainly rely on accuracy. Of course, in spit of the facts behind the precision, recall and f-measure, availability of their values would be helpful when comparing those studies which have just reported the evaluation results for them, rather than accuracy.

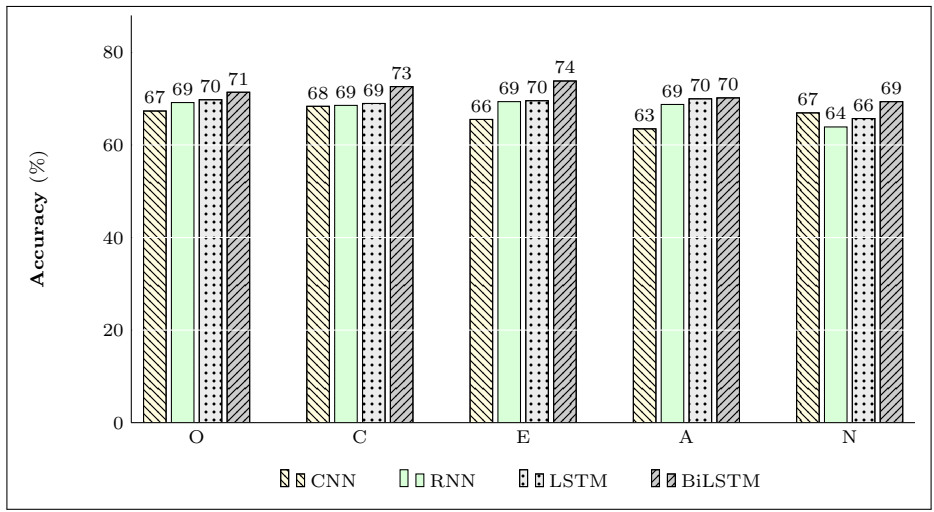
\begin{figure}[htbp]
	
	\scriptsize
	\makebox[\textwidth]
	{		
		\fbox
		{
			\begin{tikzpicture}
				\centering
				\begin{axis}[
					ybar, axis on top,
					height=8cm, width=12cm,
					bar width=0.3cm,
					ymajorgrids, tick align=inside,
					major grid style={draw=white},
					enlarge y limits={value=.1,upper},
					ymin=0, ymax=80,
					axis x line*=bottom,
					axis y line*=left,
					width=12cm,   
					height=7cm,   
					y axis line style={opacity=100},
					tickwidth=0pt,
					enlarge x limits=true,
					legend style={
						draw=none,   
						at={(0.5,-0.1)},
						legend cell align=left,
						anchor=north,
						legend columns=4,	
						/tikz/every even column/.append style={column sep=0.5cm}
					},
					ylabel={\textbf{Accuracy} (\%)},
					symbolic x coords={
						O,C,E,A,N},
					xtick=data,
					nodes near coords={
						\pgfmathprintnumber[precision=0]{\pgfplotspointmeta}
					}
					]
					
					\addplot [black,fill=yellow!15, postaction={pattern=north west lines}] coordinates {	
						(O, 67.34)
						(C, 68.36) 
						(E, 65.52)
						(A, 63.49) 
						(N, 66.94) };
					\addplot [black,fill=green!15] coordinates {	
						(O, 69.17)
						(C, 68.56) 
						(E, 69.37)
						(A, 68.76) 
						(N, 63.90)};
					\addplot [black, fill=gray!15, postaction={pattern=dots}] coordinates {	
						(O, 69.78)
						(C, 68.97) 
						(E, 69.57)
						(A, 69.98) 
						(N, 65.72)};
					\addplot [black, fill=gray!40 , postaction={pattern= north east lines}] coordinates {	
						(O, 71.40)
						(C, 72.62) 
						(E, 73.83)
						(A, 70.18) 
						(N, 69.37)};

					\legend{CNN, RNN, LSTM, BiLSTM}
					
				\end{axis}
			\end{tikzpicture}
		}
	}
	\caption{\protect\raggedright Accuracy values for four suggested APP classifiers, in each of the five personality traits in Big Five model (results are rounded)}
	
	\label{fig_OCEAN_Accuracies}		
\end{figure}

\hyperref[table_Evaluation_results_Table]{Table~\ref*{table_Evaluation_results_Table}}, provides the results obtained from the evaluation of four APP classifiers. Comparing accuracy values among four suggested classifiers, the most striking results were achieved through BiLSTM. Specifically, it had the most accurate predictions in all \textit{OCEAN} traits compared to other classifiers. Hence, the first highest average accuracy in five traits was achieved by BiLSTM. In addition, the second highest average accuracy was attained by LSTM. However, comparing the accuracies in each trait individually, reveals that it had more accurate predictions in \textit{O}, \textit{C}, \textit{E}, and \textit{A} rather than RNN and CNN; while CNN in \textit{N} practically had better predictions. However, LSTM concluded more accurate results rather than simple RNN in \textit{N}. Afterwards RNN outperformed CNN in all traits except \textit{N}. \hyperref[fig_OCEAN_Accuracies]{Figure~\ref*{fig_OCEAN_Accuracies}}, compares the accuracy values in five personality traits resulted from four classification models.

As the \hyperref[table_Evaluation_results_Table]{Table~\ref*{table_Evaluation_results_Table}} shows, the same ranking as it happens when perusing the accuracy, was achieved by classification models when taking the average f-measure into consideration. That is to say, BiLSTM, LSTM, RNN and at last CNN were ranked first to fourth, respectively. Although, the ranks do not last when comparing the f-measure values individually in each trait. Regarding recall average values, LSTM with a slight difference to BiLSTM and CNN, had better performance. At last among four suggested classifiers, BiLSTM, LSTM, RNN and CNN had the most precise predictions respectively, as it can be seen from the \hyperref[table_Evaluation_results_Table]{Table~\ref*{table_Evaluation_results_Table}}.

\begin{table}
	\centering
	\caption{Evaluation results for suggested APP classifiers, including CNN-based, RNN-based, LSTM-based, and BiLSTM-based classifiers}
	\label{table_Evaluation_results_Table}
	\scriptsize
	\begin{adjustbox}{max width=\textwidth}
		\begin{tabular}{|c|c|ccccc|c|} 
			\hline
			\rowcolor{gray!40} 		
			\textbf{Metric} & \textbf{Classification Model} & \textbf{O} & \textbf{C} & \textbf{E} & \textbf{A} & \textbf{N} & \textbf{Avg.}\\ 
			\hline
			\hline
			\multirow{4}*{\textbf{Precision}}
			& \cellcolor{gray!25} CNN & \cellcolor{gray!25} 61.22 & \cellcolor{gray!25} 60.70 &	\cellcolor{gray!25} 60.13 & \cellcolor{gray!25} 59.80 & \cellcolor{gray!25} 62.42 & \cellcolor{gray!25} 60.85 \\ 
			
			& \cellcolor{yellow!15} RNN  & \cellcolor{yellow!15} 62.41 & \cellcolor{yellow!15} 62.80 & \cellcolor{yellow!15} 64.49 & \cellcolor{yellow!15} 61.63 & \cellcolor{yellow!15} 55.52 & \cellcolor{yellow!15} 61.37 \\ 
			
			& \cellcolor{gray!25} LSTM & \cellcolor{gray!25} 66.56 & \cellcolor{gray!25} 67.65 & \cellcolor{gray!25} 63.79 & \cellcolor{gray!25} 67.58 & \cellcolor{gray!25} 59.62 & \cellcolor{gray!25} 65.04 \\ 
			
			& \cellcolor{yellow!15} BiLSTM & \cellcolor{yellow!15} \textbf{69.12} & \cellcolor{yellow!15} \textbf{71.43} & \cellcolor{yellow!15} \textbf{73.05} & \cellcolor{yellow!15} \textbf{67.75} & \cellcolor{yellow!15} \textbf{62.69} & \cellcolor{yellow!15} \textbf{68.81} \\
			
			\hline
			\hline
			
			\multirow{4}*{\textbf{Recall}}
			& \cellcolor{gray!25} CNN & \cellcolor{gray!25} \textbf{82.68} & \cellcolor{gray!25} 79.72 & \cellcolor{gray!25} 78.36 & \cellcolor{gray!25} 74.37 & \cellcolor{gray!25} \textbf{78.48} & \cellcolor{gray!25} 78.72 \\ 
			
			& \cellcolor{yellow!15} RNN  & \cellcolor{yellow!15} 77.73 & \cellcolor{yellow!15} 71.69 & \cellcolor{yellow!15} 77.06 & \cellcolor{yellow!15} 74.30 & \cellcolor{yellow!15} 76.67 & \cellcolor{yellow!15} 75.49 \\ 
			
			& \cellcolor{gray!25} LSTM & \cellcolor{gray!25} 81.20 & \cellcolor{gray!25} 79.31 & \cellcolor{gray!25} 82.76 & \cellcolor{gray!25} \textbf{78.88} & \cellcolor{gray!25} 71.82 & \cellcolor{gray!25} \textbf{78.79} \\ 
			
			& \cellcolor{yellow!15} BiLSTM & \cellcolor{yellow!15} 78.80 & \cellcolor{yellow!15} \textbf{80.46} & \cellcolor{yellow!15} \textbf{83.03} & \cellcolor{yellow!15} 76.33 & \cellcolor{yellow!15} 75.12 & \cellcolor{yellow!15} 78.75 \\ 
			
			\hline
			\hline
			
			\multirow{4}*{\textbf{F-measure}}
			& \cellcolor{gray!25} CNN & \cellcolor{gray!25} 70.35 & \cellcolor{gray!25} 68.92 & \cellcolor{gray!25} 68.05 & \cellcolor{gray!25} 66.29 & \cellcolor{gray!25} \textbf{69.53} & \cellcolor{gray!25} 68.63 \\ 
			
			& \cellcolor{yellow!15} RNN  & \cellcolor{yellow!15} 69.23 & \cellcolor{yellow!15} 66.95 & \cellcolor{yellow!15} 70.22 & \cellcolor{yellow!15} 67.37 & \cellcolor{yellow!15} 64.40 & \cellcolor{yellow!15} 67.64 \\ 
			
			& \cellcolor{gray!25} LSTM & \cellcolor{gray!25} 73.15 & \cellcolor{gray!25} 73.02 & \cellcolor{gray!25} 72.05 & \cellcolor{gray!25} \textbf{72.79} & \cellcolor{gray!25} 65.15 & \cellcolor{gray!25} 71.23 \\ 
			
			& \cellcolor{yellow!15} BiLSTM & \cellcolor{yellow!15} \textbf{73.64} & \cellcolor{yellow!15} \textbf{75.68} & \cellcolor{yellow!15} \textbf{77.72} & \cellcolor{yellow!15} 71.78 & \cellcolor{yellow!15} 68.34 & \cellcolor{yellow!15} \textbf{73.43} \\ 
		
			\hline
			\hline
			
			\multirow{4}*{\textbf{Accuracy}}
			& CNN & 67.34 & 68.36 & 65.52 & 63.49 & 66.94 & 66.33 \\ 
			& \cellcolor{green!15} RNN  & \cellcolor{green!15} 69.17 & \cellcolor{green!15} 68.56 & \cellcolor{green!15} 69.37 & \cellcolor{green!15} 68.76 & \cellcolor{green!15} 63.90 & \cellcolor{green!15} 67.95 \\ 
			& LSTM & 69.78 & 68.97 & 69.57 & 69.98 & 65.72 & 68.44 \\ 
			& \cellcolor{green!15} BiLSTM & \cellcolor{green!15} \textbf{71.40} & \cellcolor{green!15} \textbf{72.62} & \cellcolor{green!15} \textbf{73.83} & \cellcolor{green!15} \textbf{70.18} & \cellcolor{green!15} \textbf{69.37} & \cellcolor{green!15} \textbf{71.48} \\ 
		
			\hline

		\end{tabular}
	\end{adjustbox}
\end{table}

\section{Discussion}
\label{sec:Disscussion}
The major objective of current study was to investigate the efficacy of \textit{knowledge graph-enabled} automatic personalty prediction system. Thus, we used four simple base deep learning classifiers which were designed maximally similar to each other. Furthermore, we intentionally avoided designing complex networks to merely appraise the efficacy of knowledge graph-enabling of an APP system. Accordingly, we suggested one CNN-based classifier as well as three recurrent classifiers; in particular a simple RNN-based classifier along with one LTSM-based and one bidirectional LSTM (BiLSTM)-based classifiers.

Regarding the resulted accuracy values in \hyperref[table_Evaluation_results_Table]{Table~\ref*{table_Evaluation_results_Table}} for each of the classifiers, it is clear that generally recurrent classifiers lead to better results, rather than CNN. It seems possible that these results are due to their ability in processing temporal information that is presented in input sequences. To put it simply, recurrent networks are 
basically designed for sequence prediction problems like text. More specifically, they are able to capture sequential information which pinpoints the existed dependencies among the words throughout the input sequence of words. 

Among three suggested recurrent networks superior results are seen for BiLSTM. In fact, it outperforms LSTM and simple RNN, which this does seem to because of that it actually is an enhanced version of LSTM, which it itself is an enhanced version of simple RNN. To rephrase it, LSTMs were proposed to tackle RNNs' problem in preserving information over several timesteps; and BiLSTMs were also proposed to tackle LSTMs' problem in ignoring future information for a given word in input sequence. In Consequence, it is fair that BiLSTMs show better results than LSTMs and simple RNNs, and LSTMs show better results than simple RNNs. What were happened in all personality traits, even though with slight deferences in some traits. So, the obtained results confirm the expectations.

Besides, CNN-based classifier leads to comparable results. Despite the fact that it was ranked forth among four classifiers, its results are so close to some recurrent classifiers in \textit{C} and even it outperforms simple RNN and LSTM-based classifiers in \textit{N}. We speculate that this might be due to its filters' good ability in feature extraction from input embedding matrices. 

The results obtained from the four proposed classifiers can be compared with the state-of-the-art APP systems which were performed on Essays Dataset in \hyperref[table_Comparison_with_StateOfTheArt]{Table~\ref*{table_Comparison_with_StateOfTheArt}}. These results go beyond previous contributions, showing that all of the suggested methods give clearly better results than all of them. On the other hand, whereas our first ranked proposed method considerably yields better results, the forth ranked proposed method also outperforms previous reports. This is an important finding in the understanding of the knowledge graph-enabling of an automatic personality prediction system. Moreover, it is anticipated that utilization of more complex classification models (like hybrid models) would lead to more accurate predictions.

\begin{table}
	\centering
	\caption{Comparing the results obtained from our proposed methods and state-of-the-art reports in APP from text, which were performed on Essays Dataset}
	\label{table_Comparison_with_StateOfTheArt}
	\scriptsize
	\begin{adjustbox}{max width=\textwidth}
		\begin{tabular}{|c||c c c c c | c||c c c c c | c|}
			\hline
			
			\multirow{2}{2.5cm} {\centering \textbf{APP Method}}& \multicolumn{6}{p{7cm}||}{\centering \cellcolor{gray!40}  \textbf{F-measure}} & \multicolumn{6}{p{7cm}|}{\centering \cellcolor{gray!40} \textbf{Accuracy}}\\
			\cline{2-13} & \multicolumn{1}{c}{\textbf{O}} & \multicolumn{1}{c}{\textbf{C}} & \multicolumn{1}{c}{\textbf{E}} & \multicolumn{1}{c}{\textbf{A}} & \multicolumn{1}{c|}{\textbf{N}} & \multicolumn{1}{c||}{\textbf{Avg.}} & \multicolumn{1}{c}{\textbf{O}} & \multicolumn{1}{c}{\textbf{C}} & \multicolumn{1}{c}{\textbf{E}} & \multicolumn{1}{c}{\textbf{A}} & \multicolumn{1}{c|}{\textbf{N}} & \multicolumn{1}{c|}{\textbf{Avg.}}\\  
			\hline
			\hline
			
			\rowcolor{green!15}
			CNN-based classifier (proposed method) & 70.35 & 68.92 & 68.05  & 66.29 & \textbf{69.53} & 68.63 & 67.34 & 68.36  & 65.52 & 63.49 & 66.94 & 66.33\\
			
			RNN-based classifier (proposed method) & 69.23 & 66.95 & 70.22  & 67.37 & 64.40 & 67.64 & 69.17 & 68.56  & 69.37 & 68.76 & 63.90 & 67.95\\
			
			\rowcolor{green!15}
			LSTM-based classifier (proposed method) & 73.15 & 73.02 & 72.05  & \textbf{72.79} & 65.15 & 71.23 & 69.78 & 68.97  & 69.57 & 69.98 & 65.72 & 68.44\\
			
			BiLSTM-based classifier (proposed method) & \textbf{73.64} & \textbf{75.68} & \textbf{77.72}  & 71.78 & 68.34 & \textbf{73.43} & \textbf{71.40} & \textbf{72.62}  & \textbf{73.83} & \textbf{70.18} & \textbf{69.37} & \textbf{71.48}\\
			\hline
			\hline
			
			\rowcolor{yellow!15}
						
			Majumder et al. \cite{Majumder2017} &  &  &   &  &  &  & 62.68 & 57.30  & 58.09 & 56.71 & 59.38 & 58.83\\
			
			\rowcolor{gray!25}
			
			Yuan et al. \cite{YuanCu2018} &  &  &   &  &  &  & 62.00 & 57.00 & 58.00 & 56.00 & 59.00 & 58.40\\
			
			\rowcolor{yellow!15}
			
			Ramezani et al. \cite{ramezani2021automatic} & 57.37 & 59.74 & 65.80  & 61.62 & 60.69 & 61.04 & 56.30 & 59.18  & 64.25 & 60.31 & 61.14 & 60.24\\
			
			\rowcolor{gray!25}
						
			Xue et al. \cite{xue2021semantic} & 67.84 & 63.46 & 71.50  & 71.92 & 62.36 & 67.42 & 63.16 & 57.49 & 58.91 & 57.49 & 59.51 & 59.31\\
						
			\rowcolor{yellow!15}
			
			El-Demerdash et al. \cite{ELDEMERDASH2020} &  &  &  &  &  &  & 63.30 & 57.97  & 58.85 & 59.25 & 59.88 & 59.85\\
			
			\rowcolor{gray!25}
			
			Jiang et al. \cite{Jiang2020} &  &  &  &  &  &  & 65.86 & 58.55  & 60.62 & 59.72 & 61.04 & 61.16\\
			
			\rowcolor{yellow!15}
			
			El-Demerdash et al. \cite{ELDASH2021} &  &  &   &  &  &  & 65.60 & 59.52 & 61.15 & 60.80 & 62.20 & 61.85\\
			
			\rowcolor{gray!25}			
			
		    Kazameini et al. \cite{kazameini2020} &  &  &   &  &  &  & 62.09 & 57.84  & 59.30 & 56.52 & 59.39 & 59.03\\
			
			\rowcolor{yellow!15}
			
			Wang et al. \cite{Wang2020Encoding} & 67.00 & 68.00 & 67.00  & 69.00 & 69.00 & 68.00 & 64.80 & 59.10  & 60.00 & 57.70 & 63.00 & 60.92\\
			
			\rowcolor{gray!25}
								
			Tighe et al. \cite{tighe2016personality} & 61.90 & 56.00 & 55.60  & 55.70 & 58.30 & 57.50 & 61.95 & 56.04  & 55.75 & 57.54 & 58.31 & 57.92\\

			\hline
		\end{tabular}

	\end{adjustbox}
\end{table}

Ultimately, we are going to answer the research questions (as stated in \hyperref[section-1]{Introduction}) according to our observations, as follows:

\textbf{RQ.1} The results of the experiment found clear support for knowledge graph-enabling of an APP system. Actually, it empower an APP system to yield considerably more accurate results. It is also worth noting that, in this study we have just utilized the embeddings of resulted knowledge graphs to perform personality predictions, while the knowledge graphs inherently comprise miscellaneous knowledges of concepts which may be effectively utilizable in automatic personality prediction.

\textbf{RQ.2} The most interesting finding was that in classification of knowledge graphs' embedding matrices, all of the proposed deep learning classifiers, namely CNN-based, RNN-based, LSTM-based, and BiLSTM-based classifiers, substantially outperform the state-of-the-art contributions in APP, in spite of the models' simple design. This is obviously confirmed when comparing our results to those of older studies. Besides, experimental observations demonstrated that the classifiers which are based on BiLSTM, LSTM, simple RNN, and CNN yield better results respectively, when they were utilized in classification of knowledge graphs' embeddings.

\textbf{RQ.3} Regarding the obtained results from several classifiers, it is clear that knowledge graph-enabling of an APP  system, totally enhances the number of accurate predictions in all personality traits of Big Five model, albeit the enhancement pattern are not similar in all of the classifiers.

\section{Conclusion}
\label{sec:Conclusion}
The current study aimed to determine the effect of knowledge graph-enabling on an automatic personality prediction system. To do so, a three-phase approach was proposed in which a given text, performs some pre-processings (including tokenization, noise removal, normalization, and named entity recognition) in its first phase. The second phase is aimed toward achieving a knowledgeable representation of input text, tries to build the corresponding knowledge graph, and then enriching it (utilizing DBpedia ontology, NRC Emotion Intensity Lexicon, and MRC psycholinguistic database), and finally embedding the enriched knowledge graph. At last in the third phase, the embedding knowledge graph is fed into some base deep learning models (namely CNN-based, simple RNN-based, LSTM-based, and BiLSTM-based classifiers) to perform personality prediction. The results demonstrate a strong effect of knowledge graph-enabling on an automatic personality prediction system. More specifically, the findings definitely confirmed the proposed method's ability to predict all five personality traits of the Big Five model.

One of the greatest practical significance of this study is that, it provides the basis of human-like behavior for machines in a specific task, namely automatic personality prediction. Since human intelligent behavior is a consequence of his/her cognitive abilities which itself is an outcome of representation of the knowledge of the world's concepts, therefore our method will help machines to mimic human behavior as it is, which is a big step forward. That is to say, providing a comprehensive representation of appearing concepts in the input text, models the human cognition for the machine which enables it to show human-like performance. The obtained results, as well as the comparison of the findings with those of other studies, confirmed this claim.

In future work, we intend to investigate more complex deep learning models, to achieve more accurate predictions. The current study has only examined the efficacy of a knowledge graph-enabled automatic personality prediction system, and hence to minimize the effect of extrinsic factors as far as possible, it just relied on simple base deep learning models. As well, since the resulted knowledge graph usually is very large, more research is also needed to find a way to cope with it. Besides, further experimental investigations are needed to peruse other graph embedding methods and determine their effectiveness. Moreover, the application of the suggested method over different datasets in different personality models could shed more light on the efficacy of the proposed method. More broadly, the proposed knowledge representation method potentially is capable of performing other tasks which deal with text, since it provides a more knowledgeable representation of text elements for machines. Hence, the issue of knowledge representation is an intriguing one which could be usefully explored in several researches.

\section*{Data Availability}
\addcontentsline{toc}{section}{Data Availability}
The data used to support the findings of this study are available from the corresponding author upon request.

\section*{Conflicts of Interest}
\addcontentsline{toc}{section}{Conflicts of Interest}
The authors declare that they have no conflicts of interest.

\addcontentsline{toc}{section}{\refname}
\bibliographystyle{unsrt}\scriptsize
\bibliography{KR_Based_APP_Bibliography.bib}

\begin{thebibliography}{10}

\bibitem{BERGNER2020100759}
Raymond~M. Bergner.
\newblock What is personality? two myths and a definition.
\newblock {\em New Ideas in Psychology}, 57:100759, 2020.

\bibitem{mairesse2007using}
Fran{\c{c}}ois Mairesse, Marilyn~A Walker, Matthias~R Mehl, and Roger~K Moore.
\newblock Using linguistic cues for the automatic recognition of personality in
  conversation and text.
\newblock {\em Journal of artificial intelligence research}, 30:457--500, 2007.

\bibitem{6113107}
Jennifer Golbeck, Cristina Robles, Michon Edmondson, and Karen Turner.
\newblock Predicting personality from twitter.
\newblock In {\em 2011 IEEE Third International Conference on Privacy,
  Security, Risk and Trust and 2011 IEEE Third International Conference on
  Social Computing}, pages 149--156, Oct 2011.

\bibitem{6406767}
Chris Sumner, Alison Byers, Rachel Boochever, and Gregory~J. Park.
\newblock Predicting dark triad personality traits from twitter usage and a
  linguistic analysis of tweets.
\newblock In {\em 2012 11th International Conference on Machine Learning and
  Applications}, volume~2, pages 386--393, Dec 2012.

\bibitem{2018Yuan}
Yahui Yuan, Baobin Li, Dongdong Jiao, and Tingshao Zhu.
\newblock The personality analysis of characters in vernacular novels by
  sc-liwc.
\newblock In Qiaohong Zu and Bo~Hu, editors, {\em Human Centered Computing},
  pages 400--409, Cham, 2018. Springer International Publishing.

\bibitem{Majumder2017}
Navonil Majumder, Soujanya Poria, Alexander Gelbukh, and Erik Cambria.
\newblock Deep learning-based document modeling for personality detection from
  text.
\newblock {\em IEEE Intelligent Systems}, 32(2):74--79, Mar 2017.

\bibitem{Sewwandi2017}
Dilini Sewwandi, Kusal Perera, Sajith Sandaruwan, Oshani Lakchani, Anupiya
  Nugaliyadde, and Samantha Thelijjagoda.
\newblock Linguistic features based personality recognition using social media
  data.
\newblock In {\em 2017 6th National Conference on Technology and Management
  (NCTM)}, pages 63--68, Jan 2017.

\bibitem{daSilva2018}
Barbara Barbosa~Claudino da~Silva and Ivandr{\'e} Paraboni.
\newblock Personality recognition from facebook text.
\newblock In Aline Villavicencio, Viviane Moreira, Alberto Abad, Helena Caseli,
  Pablo Gamallo, Carlos Ramisch, Hugo Gon{\c{c}}alo~Oliveira, and
  Gustavo~Henrique Paetzold, editors, {\em Computational Processing of the
  Portuguese Language}, pages 107--114, Cham, 2018. Springer International
  Publishing.

\bibitem{YuanCu2018}
Cuixin Yuan, Junjie Wu, Hong Li, and Lihong Wang.
\newblock Personality recognition based on user generated content.
\newblock In {\em 2018 15th International Conference on Service Systems and
  Service Management (ICSSSM)}, pages 1--6, July 2018.

\bibitem{YuJianguo2017}
Jianguo Yu and Konstantin Markov.
\newblock Deep learning based personality recognition from facebook status
  updates.
\newblock In {\em 2017 IEEE 8th International Conference on Awareness Science
  and Technology (iCAST)}, pages 383--387, Nov 2017.

\bibitem{SunXian2018}
Xiangguo Sun, Bo~Liu, Jiuxin Cao, Junzhou Luo, and Xiaojun Shen.
\newblock Who am i? personality detection based on deep learning for texts.
\newblock In {\em 2018 IEEE International Conference on Communications (ICC)},
  pages 1--6, May 2018.

\bibitem{yilmazTun2019}
Tuncay Y{\i}lmaz, Abdullah Ergil, and Bahar {\.{I}}lgen.
\newblock Deep learning-based document modeling for personality detection from
  turkish texts.
\newblock In Kohei Arai, Rahul Bhatia, and Supriya Kapoor, editors, {\em
  Proceedings of the Future Technologies Conference (FTC) 2019}, pages
  729--736, Cham, 2020. Springer International Publishing.

\bibitem{ELDASH2021}
Kamal El-Demerdash, Reda~A. El-Khoribi, Mahmoud~A. {Ismail Shoman}, and Sherif
  Abdou.
\newblock Deep learning based fusion strategies for personality prediction.
\newblock {\em Egyptian Informatics Journal}, 2021.

\bibitem{REN2021102532}
Zhancheng Ren, Qiang Shen, Xiaolei Diao, and Hao Xu.
\newblock A sentiment-aware deep learning approach for personality detection
  from text.
\newblock {\em Information Processing \& Management}, 58(3):102532, 2021.

\bibitem{xue2018deep}
Di~Xue, Lifa Wu, Zheng Hong, Shize Guo, Liang Gao, Zhiyong Wu, Xiaofeng Zhong,
  and Jianshan Sun.
\newblock Deep learning-based personality recognition from text posts of online
  social networks.
\newblock {\em Applied Intelligence}, 48(11):4232--4246, 2018.

\bibitem{JEREMY2021416}
Nicholaus~Hendrik Jeremy and Derwin Suhartono.
\newblock Automatic personality prediction from indonesian user on twitter
  using word embedding and neural networks.
\newblock {\em Procedia Computer Science}, 179:416--422, 2021.
\newblock 5th International Conference on Computer Science and Computational
  Intelligence 2020.

\bibitem{christian2021text}
Hans Christian, Derwin Suhartono, Andry Chowanda, and Kamal~Z Zamli.
\newblock Text based personality prediction from multiple social media data
  sources using pre-trained language model and model averaging.
\newblock {\em Journal of Big Data}, 8(1):1--20, 2021.

\bibitem{bergman2018knowledge}
Michael~K Bergman, Michael~K Bergman, and Lagerstrom-Fife.
\newblock {\em Knowledge Representation Practionary}.
\newblock Springer, 2018.

\bibitem{van2008handbook}
Frank Van~Harmelen, Vladimir Lifschitz, and Bruce Porter.
\newblock {\em Handbook of knowledge representation}.
\newblock Elsevier, 2008.

\bibitem{Matz2016}
Sandra Matz, Yin Wah~Fiona Chan, and Michal Kosinski.
\newblock {\em Models of Personality}, pages 35--54.
\newblock Springer International Publishing, Cham, 2016.

\bibitem{BigFive1992}
Robert~R McCrae and Oliver~P John.
\newblock An introduction to the five-factor model and its applications.
\newblock {\em Journal of personality}, 60(2):175--215, 1992.

\bibitem{Moreno2021}
José~David Moreno, José~Á. Martínez-Huertas, Ricardo Olmos, Guillermo
  Jorge-Botana, and Juan Botella.
\newblock Can personality traits be measured analyzing written language? a
  meta-analytic study on computational methods.
\newblock {\em Personality and Individual Differences}, 177:110818, 2021.

\bibitem{2019introductionToPsych}
Jorden~A Cummings and Lee Sanders.
\newblock {\em Introduction to Psychology}.
\newblock University of Saskatchewan Open Press, 2019.

\bibitem{ramezani2021automatic}
Majid Ramezani, Mohammad-Reza Feizi-Derakhshi, Mohammad-Ali Balafar, Meysam
  Asgari-Chenaghlu, Ali-Reza Feizi-Derakhshi, Narjes Nikzad-Khasmakhi, Mehrdad
  Ranjbar-Khadivi, Zoleikha Jahanbakhsh-Nagadeh, Elnaz Zafarani-Moattar, and
  Taymaz Akan.
\newblock Automatic personality prediction: an enhanced method using ensemble
  modeling.
\newblock {\em Neural Computing and Applications}, Jun 2022.

\bibitem{feizi2021state}
Ali-Reza Feizi-Derakhshi, Mohammad-Reza Feizi-Derakhshi, Majid Ramezani, Narjes
  Nikzad-Khasmakhi, Meysam Asgari-Chenaghlu, Taymaz Akan, Mehrdad
  Ranjbar-Khadivi, Elnaz Zafarni-Moattar, and Zoleikha Jahanbakhsh-Naghadeh.
\newblock The state-of-the-art in text-based automatic personality prediction,
  2021.

\bibitem{MORENO2021110818}
José~David Moreno, José~Á. Martínez-Huertas, Ricardo Olmos, Guillermo
  Jorge-Botana, and Juan Botella.
\newblock Can personality traits be measured analyzing written language? a
  meta-analytic study on computational methods.
\newblock {\em Personality and Individual Differences}, 177:110818, 2021.

\bibitem{HAN2020105550}
Songqiao Han, Hailiang Huang, and Yuqing Tang.
\newblock Knowledge of words: An interpretable approach for personality
  recognition from social media.
\newblock {\em Knowledge-Based Systems}, 194:105550, 2020.

\bibitem{pennebaker2001liwc}
James~W Pennebaker, Martha~E Francis, and Roger~J Booth.
\newblock Linguistic inquiry and word count: Liwc 2001.
\newblock {\em Mahway: Lawrence Erlbaum Associates}, 71(2001):2001, 2001.

\bibitem{xue2021semantic}
Xia Xue, Jun Feng, and Xia Sun.
\newblock Semantic-enhanced sequential modeling for personality trait
  recognition from texts.
\newblock {\em Applied Intelligence}, pages 1--13, 2021.

\bibitem{MehtaYash2020}
Yash Mehta, Samin Fatehi, Amirmohammad Kazameini, Clemens Stachl, Erik Cambria,
  and Sauleh Eetemadi.
\newblock Bottom-up and top-down: Predicting personality with psycholinguistic
  and language model features.
\newblock In {\em 2020 IEEE International Conference on Data Mining (ICDM)},
  pages 1184--1189, Nov 2020.

\bibitem{ELDEMERDASH2020}
Kamal El-Demerdash, Reda~A. El-Khoribi, Mahmoud A.~Ismail Shoman, and Sherif
  Abdou.
\newblock Psychological human traits detection based on universal language
  modeling.
\newblock {\em Egyptian Informatics Journal}, 2020.

\bibitem{Jiang2020}
Hang Jiang, Xianzhe Zhang, and Jinho~D. Choi.
\newblock Automatic text-based personality recognition on monologues and
  multiparty dialogues using attentive networks and contextual embeddings
  (student abstract).
\newblock {\em Proceedings of the AAAI Conference on Artificial Intelligence},
  34(10):13821--13822, Apr. 2020.

\bibitem{Kunte2020}
Aditi Kunte and Suja Panicker.
\newblock Personality prediction of social network users using ensemble and
  xgboost.
\newblock In Himansu Das, Prasant~Kumar Pattnaik, Siddharth~Swarup Rautaray,
  and Kuan-Ching Li, editors, {\em Progress in Computing, Analytics and
  Networking}, pages 133--140, Singapore, 2020. Springer Singapore.

\bibitem{Ashima2019ensembel}
Ashima Sood and Rekha Bhatia.
\newblock Baron-cohen model based personality classification using ensemble
  learning.
\newblock In Durgesh~Kumar Mishra, Xin-She Yang, and Aynur Unal, editors, {\em
  Data Science and Big Data Analytics}, pages 57--65, Singapore, 2019. Springer
  Singapore.

\bibitem{kazameini2020}
Amirmohammad Kazameini, Samin Fatehi, Yash Mehta, Sauleh Eetemadi, and Erik
  Cambria.
\newblock Personality trait detection using bagged svm over bert word embedding
  ensembles.
\newblock {\em arXiv preprint arXiv:2010.01309}, 2020.

\bibitem{sun2019group}
Xiangguo Sun, Bo~Liu, Qing Meng, Jiuxin Cao, Junzhou Luo, and Hongzhi Yin.
\newblock Group-level personality detection based on text generated networks.
\newblock {\em World Wide Web}, 23(3):1--20, 2019.

\bibitem{9172426}
Zhanming Guan, Bin Wu, Bai Wang, and Hezi Liu.
\newblock Personality2vec: Network representation learning for personality.
\newblock In {\em 2020 IEEE Fifth International Conference on Data Science in
  Cyberspace (DSC)}, pages 30--37, July 2020.

\bibitem{pennebaker1999linguistic}
James~W Pennebaker and Laura~A King.
\newblock Linguistic styles: language use as an individual difference.
\newblock {\em Journal of personality and social psychology}, 77(6):1296, 1999.

\bibitem{6876017}
Alexander Lex, Nils Gehlenborg, Hendrik Strobelt, Romain Vuillemot, and
  Hanspeter Pfister.
\newblock Upset: Visualization of intersecting sets.
\newblock {\em IEEE Transactions on Visualization and Computer Graphics},
  20(12):1983--1992, 2014.

\bibitem{schutze2008introduction}
Hinrich Sch{\"u}tze, Christopher~D Manning, and Prabhakar Raghavan.
\newblock {\em Introduction to information retrieval}, volume~39.
\newblock Cambridge University Press Cambridge, 2008.

\bibitem{bird2009natural}
Steven Bird, Ewan Klein, and Edward Loper.
\newblock {\em Natural language processing with Python: analyzing text with the
  natural language toolkit}.
\newblock " O'Reilly Media, Inc.", 2009.

\bibitem{spacy}
Matthew Honnibal, Ines Montani, Sofie Van~Landeghem, and Adriane Boyd.
\newblock {spaCy: Industrial-strength Natural Language Processing in Python},
  2020.

\bibitem{DBpedia2007}
S{\"o}ren Auer, Christian Bizer, Georgi Kobilarov, Jens Lehmann, Richard
  Cyganiak, and Zachary Ives.
\newblock Dbpedia: A nucleus for a web of open data.
\newblock In Karl Aberer, Key-Sun Choi, Natasha Noy, Dean Allemang, Kyung-Il
  Lee, Lyndon Nixon, Jennifer Golbeck, Peter Mika, Diana Maynard, Riichiro
  Mizoguchi, Guus Schreiber, and Philippe Cudr{\'e}-Mauroux, editors, {\em The
  Semantic Web}, pages 722--735, Berlin, Heidelberg, 2007. Springer Berlin
  Heidelberg.

\bibitem{CHEN2020KG}
Xiaojun Chen, Shengbin Jia, and Yang Xiang.
\newblock A review: Knowledge reasoning over knowledge graph.
\newblock {\em Expert Systems with Applications}, 141:112948, 2020.

\bibitem{ChenZhe2020}
Zhe Chen, Yuehan Wang, Bin Zhao, Jing Cheng, Xin Zhao, and Zongtao Duan.
\newblock Knowledge graph completion: A review, 2020.

\bibitem{StanfordKG2021}
Denny Vrandecic, Jans Aasman, and Mikhail Galkin.
\newblock What is a knowledge graph?, 2020.

\bibitem{chein2008graph}
Michel Chein and Marie-Laure Mugnier.
\newblock {\em Graph-based knowledge representation: computational foundations
  of conceptual graphs}.
\newblock Springer Science \& Business Media, 2008.

\bibitem{KnowledgeGraphsBook2021}
Mayank Kejriwal, Craig~A Knoblock, and Pedro Szekely.
\newblock {\em Knowledge Graphs: Fundamentals, Techniques, and Applications}.
\newblock MIT Press, 2021.

\bibitem{RDFHiba2018}
Hiba Arnaout and Shady Elbassuoni.
\newblock Effective searching of rdf knowledge graphs.
\newblock {\em Journal of Web Semantics}, 48:66--84, 2018.

\bibitem{Hogan2020}
Aidan Hogan.
\newblock {\em SPARQL Query Language}, pages 323--448.
\newblock Springer International Publishing, Cham, 2020.

\bibitem{jakus2013book}
Grega Jakus, Veljko Milutinovi{\'c}, Sanida Omerovi{\'c}, and Sa{\v{s}}o
  Toma{\v{z}}i{\v{c}}.
\newblock {\em Concepts, ontologies, and knowledge representation}.
\newblock Springer, 2013.

\bibitem{mw_ontology}
Merriam-Webster.
\newblock Ontology.
\newblock In {\em Merriam-Webster.com dictionary}.

\bibitem{DBpediaOntologyClasses}
MultiMedia LLC.
\newblock Dbpedia ontology classes, 2021.

\bibitem{NRCAFFECTDic}
Saif~M. Mohammad.
\newblock Word affect intensities.
\newblock In {\em Proceedings of the 11th Edition of the Language Resources and
  Evaluation Conference (LREC-2018)}, Miyazaki, Japan, 2018.

\bibitem{MRCdb}
Michael Wilson.
\newblock Mrc psycholinguistic database: Machine-usable dictionary, version
  2.00.
\newblock {\em Behavior research methods, instruments, \& computers},
  20(1):6--10, 1988.

\bibitem{ristoski2019rdf2vec}
Petar Ristoski, Jessica Rosati, Tommaso Di~Noia, Renato De~Leone, and Heiko
  Paulheim.
\newblock Rdf2vec: Rdf graph embeddings and their applications.
\newblock {\em Semantic Web}, 10(4):721--752, 2019.

\bibitem{mikolov2013WORD2VEC}
Tomas Mikolov, Kai Chen, Greg Corrado, and Jeffrey Dean.
\newblock Efficient estimation of word representations in vector space.
\newblock {\em arXiv preprint arXiv:1301.3781}, 2013.

\bibitem{CNN2018}
Jiuxiang Gu, Zhenhua Wang, Jason Kuen, Lianyang Ma, Amir Shahroudy, Bing Shuai,
  Ting Liu, Xingxing Wang, Gang Wang, Jianfei Cai, and Tsuhan Chen.
\newblock Recent advances in convolutional neural networks.
\newblock {\em Pattern Recognition}, 77:354--377, 2018.

\bibitem{RNN2020}
Alex Sherstinsky.
\newblock Fundamentals of recurrent neural network (rnn) and long short-term
  memory (lstm) network.
\newblock {\em Physica D: Nonlinear Phenomena}, 404:132306, 2020.

\bibitem{BiLSTM2019}
Yong Yu, Xiaosheng Si, Changhua Hu, and Jianxun Zhang.
\newblock {A Review of Recurrent Neural Networks: LSTM Cells and Network
  Architectures}.
\newblock {\em Neural Computation}, 31(7):1235--1270, 07 2019.

\bibitem{Wang2020Encoding}
Zhe Wang, Chun-Hua Wu, Qing-Biao Li, Bo~Yan, and Kang-Feng Zheng.
\newblock Encoding text information with graph convolutional networks for
  personality recognition.
\newblock {\em Applied Sciences}, 10(12), 2020.

\bibitem{tighe2016personality}
Edward~P Tighe, Jennifer~C Ureta, Bernard Andrei~L Pollo, Charibeth~K Cheng,
  and Remedios de~Dios~Bulos.
\newblock Personality trait classification of essays with the application of
  feature reduction.
\newblock In {\em SAAIP@ IJCAI}, pages 22--28, 2016.

\end{thebibliography}

\end{document}